%% file: main_arxiv.tex
\documentclass[journal,12pt,onecolumn,draftclsnofoot,]{IEEEtran}
\usepackage{times}

\usepackage[numbers]{natbib}
\usepackage{multicol}
\usepackage[bookmarks=true]{hyperref}
\usepackage{amsfonts,amsmath}
\usepackage{xcolor}
\usepackage{subcaption}
\usepackage{graphicx} 
\usepackage{cases}
\usepackage{multirow}
\usepackage[linesnumbered,ruled]{algorithm2e}

\usepackage{alphalph}

\newcommand{\etal}{\textit{et al}.}
\newcommand{\eg}{\textit{e.g.}}
\newcommand{\ie}{\textit{i.e.}}

\DeclareMathOperator*{\argmax}{arg\,max}

\newtheorem{definition}{Definition}
\newtheorem{theorem}{Theorem}

\begin{document}

\title{Effort Allocation for Deadline-Aware Task and Motion Planning: A Metareasoning Approach}

\author{Yoonchang Sung*, Shahaf S. Shperberg*, Qi Wang*, and Peter Stone
\thanks{Y. Sung is with the Department of Computer Science, The University of Texas at Austin, Austin, TX 78712 USA (e-mail:yooncs8@cs.utexas.edu).}
\thanks{S. Shperberg is with the Department of Software and Information Systems Engineering, Ben-Gurion University of the Negev, Be'er Sheva, Israel (e-mail:shperbsh@bgu.ac.il).}
\thanks{Q. Wang is with the Department of Computer Science, The University of Texas at Austin, Austin, TX 78712 USA (e-mail:harrywang@utexas.edu).}
\thanks{P. Stone is with the Department of Computer Science, The University of Texas at Austin, Austin, TX 78712 USA and Sony AI (e-mail:pstone@cs.utexas.edu).}
\thanks{*Equal contribution.}
}



\maketitle

\begin{abstract}
In robot planning, tasks can often be achieved through multiple options, each consisting of several actions. This work specifically addresses deadline constraints in task and motion planning, aiming to find a plan that can be executed within the deadline despite uncertain planning and execution times. We propose an effort allocation problem, formulated as a Markov decision process (MDP), to find such a plan by leveraging metareasoning perspectives to allocate computational resources among the given options. We formally prove the NP-hardness of the problem by reducing it from the knapsack problem.

Both a model-based approach, where transition models are learned from past experience, and a model-free approach, which overcomes the unavailability of prior data acquisition through reinforcement learning, are explored. For the model-based approach, we investigate Monte Carlo tree search (MCTS) to approximately solve the proposed MDP and further design heuristic schemes to tackle NP-hardness, leading to the approximate yet efficient algorithm called DP\_Rerun. In experiments, DP\_Rerun demonstrates promising performance comparable to MCTS while requiring negligible computation time.
\end{abstract}

\begin{IEEEkeywords}
Metareasoning, task and motion planning, resource-constrained planning.
\end{IEEEkeywords}

\section{Introduction}
\label{sec:intro}

\IEEEPARstart{I}{magine} a scenario where a bus leaves from the station in 30 minutes. There are multiple routes to the station, but it is not known exactly how long each route will take, and even the computation times for determining the exact motions needed to execute each route are unknown a priori. Similarly, imagine you are a chef and need to serve a meal in 15 minutes, which requires a pot. There is a nearby pot at the bottom of a large stack of dishes such that retrieving it involves taking out several other objects, while taking out another pot located far away does not involve taking out any other objects. In such scenarios, we need to reason about how long each option will take to execute, combined with how long it will take to even find an executable plan.

Even without deadlines, the above planning problems are challenging for robots, particularly when the planning horizon is long, as finding a solution generally involves searching in high dimensions with significant depth and branching factors. Task and motion planning (TAMP) introduces abstractions—such as passing through a particular intersection in the bus example and picking up a dish in the meal preparation example—defined at the symbolic level. These abstractions do not specify how the robot can realize low-level motions (\ie, trajectories or motor commands) to achieve these abstract actions, allowing for planning in the abstract space and avoiding the search complexity of the low-level motion space.  However, such planning is generally done without full knowledge of how long it will take to translate an abstract action into a fully executable sequence of low-level commands.

Introducing the deadline constraint to TAMP problems induces an additional problem of scheduling computation allocation among candidate abstract plans. An abstract plan, referred to as a plan skeleton, corresponds to a sequence of abstract actions. The TAMP problem and computation selection may be interleaved, with computation selection determining which abstract plan the planner must consider and affecting the probability of finding a solution within a deadline. Essentially, computation selection can be viewed as metareasoning to effectively solve deadline-constrained TAMP problems.

In this work, we introduce a Markov decision process (MDP) formulation that involves metareasoning for computation selection in deadline-constrained TAMP problems. The introduced problem is inherently stochastic, as exact motions are unknown a priori before computation, resulting in stochastic planning and execution times. To address this stochasticity, we explore both model-based and model-free approaches. The model-based approach involves learning transition models when data on past planning experience
is available, while the model-free approach is used when such data is unavailable.

We additionally show that the proposed metareasoning MDP problem is NP-hard, proven by reduction from the knapsack problem. To alleviate the complexity of the problem, we employ the Monte Carlo tree search (MCTS) scheme to approximately solve the problem and further propose heuristics that exploit the structure of an optimal policy, which can be solved in polynomial time at the expense of optimality. However, in experiments, we show that the problem restricted by heuristics yields solutions that are comparable to those from the original problem while greatly improving efficiency. 

Our main contributions can be summarized as follows:
\begin{itemize}
\item We propose a deadline-aware TAMP formulation that leverages metareasoning perspectives. To the best of our knowledge, our work is the first to address time-critical scenarios in TAMP.
\item We formally prove the NP-hardness of the proposed metareasoning problem and design a polynomial-time approximate algorithm, called DP\_Rerun, through heuristics.
\item Furthermore, we investigate approximately solving the metareasoning problem using MCTS and employing reinforcement learning when data is not available a priori.
\item We test both navigation and manipulation scenarios to evaluate the effectiveness of the proposed method, DP\_Rerun, in practice. We find empirically that it significantly outperforms baselines in experiments. 
In particular, DP\_Rerun takes significantly less computation time to achieve performance almost on par with MCTS.
\end{itemize}

The rest of this paper is organized as follows. Section~\ref{sec:related} introduces related work on TAMP and metareasoning. Section~\ref{sec:prob} presents our metareasoning formulation as an MDP problem. The hardness of the proposed problem is derived in Section~\ref{sec:hardness}. Model-based and model-free solution methods are introduced in Sections~\ref{sec:model} and \ref{sec:free}, respectively. Section~\ref{sec:example} illustrates the practical use cases of the proposed method in navigation and manipulation domains. Section~\ref{sec:exp} presents the experimental results, and concluding remarks are included in Section~\ref{sec:conc}.

\section{Related Work}
\label{sec:related}

Here, we review the related literature on task and motion planning, resource-constrained planning, and metareasoning to highlight the overlooked challenges addressed in this work.

\subsection{Task and motion planning}
\label{subsec:tamp}
TAMP problems involve manipulating multiple objects in the world, requiring reasoning about where to grasp an object, where to place it, and robot motion commands~\cite{garrett2021integrated,zhao2024survey}. TAMP planners introduce symbolic abstractions that reason primarily about objects to separate out low-level motion planning. Essentially, TAMP exhibits a bilevel structure: high-level task reasoning (\ie, determining which actions to take) and low-level motion reasoning (\ie, determining how to execute these actions) mutually provide complementary guidance. 

Existing TAMP planners can be generally classified into three approaches~\cite{garrett2021integrated}. The \emph{satisfy-before-sequence} approach~\cite{hauser2011randomized,krontiris2015dealing,garrett2018ffrob} involves finding a valid assignment of values to continuous variables that satisfy the corresponding constraints. Subsequently, the sampled values are sequenced to generate a complete plan executable by the robot. On the other hand, the \emph{sequence-before-satisfy} approach~\cite{wolfe2010combined,srivastava2014combined,toussaint2015logic,lagriffoul2016combining,dantam2018incremental,lo2020petlon,garrett2020pddlstream,sung2023learning} reverses the search process by first computing abstract plans that do not involve reasoning about continuous variables, followed by finding satisfying values for those variables. An alternative approach is \emph{interleaved-satisfy-and-sequence}~\cite{plaku2010sampling,kaelbling2013integrated,kim2022representation,kingston2022scaling}, whereby a search tree is constructed to jointly search for satisfying values for both discrete and continuous variables in an interleaved manner.

Two main directions have been actively pursued in the community to advance the state of the art in TAMP research: (1) focusing on improving planning efficiency and (2) developing a learning framework to acquire models for TAMP planners from data, rather than relying solely on hand-designed models. Several ideas have been proposed to enhance planning efficiency, such as novel heuristics for informed search control~\cite{garrett2018ffrob,kim2019learning,sung2023learning}, learning samplers for effectively handling continuous variables~\cite{chitnis2016guided,fang2023dimsam}, value function learning to estimate the contribution to reaching a goal~\cite{kim2022representation}, and feasibility prediction to mitigate the failure of expensive motion planning~\cite{wells2019learning,driess2020deep,li2023sampling,yang2023sequence,sung2023motion}. TAMP planners require models, such as symbolic models and skills, to effectively find a solution. Data-driven model learning efforts for TAMP include skill learning~\cite{wang2021learning,silver2021learning,silver2022learning,cheng2023nod,mishra2023generative}, world model learning~\cite{curtis2022long,ding2022learning}, and predicate learning~\cite{silver2023predicate}.

As such, most TAMP research has focused on efficiently and effectively finding a satisfying plan, while little has addressed optimal planning~\cite{vega2020asymptotically,shome2021pushing,thomason2022task}. To the best of our knowledge, no previous work has tackled deadline-aware planning in TAMP. This work introduces the first deadline-aware TAMP formulation, highlighting the distinctive challenges posed by stochastic planning and execution times, as well as the allocation of computation among available options.

\subsection{Resource-constrained planning}
\label{subsec:resource}

In robot planning, it is often essential to consider resource constraints to adhere to limited time or energy budgets, given that a robot's operational lifetime is finite. Various forms of these constraints have been explored in the literature~~\cite{sung2023survey}; here, we only discuss representative classes of problems. 

\emph{Prize collecting traveling salesman problem}~\cite{balas1989prize} is a variant of the well-known traveling salesman problem. Its goal is to visit a subset of vertices from a given graph in a way that minimizes both the traveling distance and the net penalties associated with collecting prizes and penalties. Another related class of problems is the \emph{orienteering problem}~\cite{chekuri2005recursive,lattimore2020bandit,liu2021team}, whose objective is to maximize the reward collected until a given budget is exhausted.
Various algorithms have been developed, including the recursive greedy algorithm~\cite{chekuri2005recursive}, which offers a provable approximate guarantee. \emph{Multi-armed bandit} problems~\cite{lattimore2020bandit} involve maximizing the expected return when dealing with a finite number of arms associated with unknown black-box reward functions. When only a fixed number of trials is allowed, resource constraints are considered, making the exploration-exploitation dilemma critical.

Although both the problems introduced in this subsection and the problem proposed in this work address bounded resources, the introduced problem includes a unique structure that makes it distinctive, thereby necessitating the consideration of metareasoning. Specifically, the problem involves two interconnected subproblems: one is the original problem of solving a given TAMP problem, and the other is the metareasoning problem of scheduling computation allocations among options to meet a deadline. We provide more detailed explanations on the connection to metareasoning in the following subsection.

\subsection{Metareasoning}
\label{subsec:metareasoning}
Rational metareasoning~\cite{DBLP:journals/ai/RussellW91} has been investigated in the context of developing intelligent agents under constraints of bounded resources, with two early proposed equivalent models: anytime algorithms~\cite{dean1988analysis} and flexible computation~\cite{horvitz1991computation}. In this subsection, our focus is on the planning-related literature. Several more comprehensive surveys can be found in the literature~\cite{cox2005metacognition,anderson2007review,ackerman2017meta,herrmann2023metareasoning}.

The \emph{value of computation} is a critical concept in metareasoning, calculated as the improvement of solution quality resulting from a computation, subtracted by the computationally incurred costs~\cite{horvitz1991computation}. Since this value is difficult to compute, how to approximate the value of computation has become a subject of research. Moreover, the improvement of solution quality in reality is often uncertain. Therefore, monitoring and control schemes have been proposed to effectively handle stochastic solution quality improvement~\cite{hansen2001monitoring,svegliato2018meta,svegliato2020model,budd24aaai}.

One constructive application of metareasoning for planning is to determine the optimal moment to stop planning and begin executing a computed plan through interleaved planning and execution~\cite{o2015metareasoning,lin2015metareasoning}. This approach is particularly valuable when the reasoning procedure exhibits anytime behavior, allowing the agent to achieve a higher-quality plan by investing more time in planning despite incurring greater costs, such as energy consumption. Therefore, monitoring the progress of plan quality improvement is crucial for establishing a stopping policy that effectively balances solution quality and computation time.

Metareasoning has also been employed in robotics problems, such as cost-effectively stopping for optimal motion planning~\cite{sung2021learning} and handling exceptional situations to ensure the safety of autonomous driving~\cite{svegliato2019belief}.

Another related constructive use case is to select which computation to utilize when multiple computations are available under limited computational resources. Metareasoning has been extensively applied to find the optimal computation among various alternatives. 
Hay~\etal~\cite{hay2012selecting} adopted metareasoning in the selection phase of MCTS to determine which future sequences to simulate and compared it with bandit settings.
Lieder~\etal~\cite{lieder2014algorithm} developed a theory for algorithm selection by metareasoning to model human strategy selection. 
Callaway~\etal~\cite{callaway2018learning} proposed a learning algorithm for approximating the optimal selection of computations. 

Our work also involves computation selection considered as metareasoning. However, there are two notable features in our work that distinguish it from the aforementioned literature. First, we address deadline-aware planning, meaning that the total duration of selected sequential computations must meet a deadline, whereas existing work treats each selected computation as a separate time-critical problem. This fundamental difference prevents the proposed work from utilizing the conventional notion of the value of computation. Second, our work considers TAMP as an object-level problem, which has not yet been addressed in the literature.


The most related work to ours is by Shperberg~\etal~\cite{DBLP:conf/aaai/ShperbergCCKRS19}, where they explored a metareasoning problem within the domain of situated temporal (symbolic) planning~\cite{DBLP:conf/aips/CashmoreCCKMR18}, a planning paradigm that considers the time elapsed during plan search while also accounting for a pre-specified deadline. They abstracted the problem of searching for plans into a meta-level scheduling problem, bypassing the complexities of plan state representation and search procedures. Their approach involved modeling the problem using a set of processes, 
each dedicated to searching for a plan, akin to representing search nodes on an open list. Each process is characterized by a probabilistic performance profile, modeled by a random variable indicating the probability of successful termination given processing time, as well as a random variable modeling the deadline corresponding to each partial plan, which is only revealed after planning is concluded. The meta-level problem lies in finding an optimal schedule of processing time across all processes that maximizes the probability that any process delivers a plan before its deadline. A simplified version of this problem, known as ``simplified allocating planning effort when actions expire,'' assumes discrete time intervals and has been proven to be NP-hard. However, under the condition of known deadlines, the problem becomes solvable in pseudo-polynomial time through dynamic programming. Later, this line of work was extended to consider interleaved planning and execution, where partial plans can be executed during the search~\cite{DBLP:conf/aaai/ElboherEtAl23,ICAPS24}.  
While this body of work bears relevance to our research, it primarily concentrates on deriving symbolic plans. In contrast, our focus lies in elaborating existing symbolic plans through motion-level reasoning to make them executable for a robot, optimizing the likelihood of meeting a pre-specified deadline.


\section{Problem Formulation}
\label{sec:prob}
Consider a robot that can interact with objects in the world. Its \emph{configuration} space is represented by $\mathbb{Q}$, and its space of \emph{grasps} is represented by $\mathbb{G}$, corresponding to a space of the 6D pose of the end-effector in $SE(3)$, the grasp preshape, and the approach direction~\cite{diankov2010automated}.

We assume that the world is composed of a finite set of objects $\mathcal{O}=\mathcal{O}_F\cup\mathcal{O}_M$, where \emph{fixed objects} $\mathcal{O}_F$ include objects such as floors, walls, tables, and shelves, and \emph{movable objects} $\mathcal{O}_M$ include objects such as cups, books, and keys that are movable by the robot. A subset of fixed objects $\mathcal{O}_F$, such as tables and shelves, is endowed with \emph{workspace regions} in $\mathbb{R}^3$, where movable objects can be placed. The space of \emph{poses} for movable object $i$ in $\mathcal{O}_M$ is denoted by $\mathbb{P}_i$, representing stable placements in the workspace regions.

The \emph{composite state space} of the robot and objects is represented by $\mathbb{Q}\times\mathbb{G}\times\prod_{i=1}^{M}\mathbb{P}_i$, where $M$ denotes the number of movable objects. Each variable corresponding to a state space component is referred to as a \emph{typed variable}, which includes a robot configuration $q\in\mathbb{Q}$, a grasp pose $g\in\mathbb{G}$, and the pose $p_i\in\mathbb{P}_i$ of each movable object $i$. The \emph{state} is then a tuple of continuous values assigned to the concatenation of these typed variables.


We assume deterministic transitions as a result of actions and error-free perception. Despite these restrictive assumptions, TAMP planning remains computationally intractable, as the underlying problems of task planning~\cite{bylander1994computational} and motion planning~\cite{reif1979complexity,canny1988complexity} have each been proven to be PSPACE-complete. Our aim is to establish an initial solution as foundational for future efforts aimed at relaxing these assumptions. However, we introduce novel challenges regarding imperfect knowledge in this work, where planning time and execution time of an action are unknown, which are crucial for addressing deadline-aware planning.

We now recap a general TAMP formulation~\cite{garrett2021integrated}, a planning framework that finds a sequence of actions to achieve a certain goal, without considering deadlines. We then broaden the problem description to include effort allocation for finding a TAMP solution that can be planned and executed within a pre-specified deadline. 

\subsection{TAMP formulation}
\label{subsec:tamp_form}
The TAMP formulation often leverages a logic-based action language, such as planning domain definition language (PDDL~\cite{fox2003pddl2}), to enable high-level symbolic planning for efficient low-level motion planning~\cite{garrett2020pddlstream}. We define a TAMP problem as a tuple $\langle\mathcal{O}, \mathcal{P}, \mathcal{I}, \mathcal{G}, \Delta\rangle$ as follows:
\begin{itemize}
\item $\mathcal{O}$ denotes a finite set of objects introduced previously.
\item $\mathcal{P}$ denotes a finite set of \emph{predicates}, where each predicate is a Boolean function that can be evaluated on a tuple of object variables $o\in\mathcal{O}$ and typed variables to determine whether it is true or false. An assignment of values to a predicate is called a \emph{literal}. For example, $\texttt{InRegion}(o_i\in\mathcal{O}_M, o_f\in\mathcal{O}_F, p_i\in\mathbb{P}_i)$ is true if the movable object $o_i$ with pose $p_i$ is in the workspace of the fixed object $o_f$, $\texttt{InHand}(o_i\in\mathcal{O}_M, g\in\mathbb{G})$ is true if the movable object $o_i$ is stably grasped by the robot with grasp pose $g$, and $\texttt{Reachable}(q\in\mathbb{Q}, q^\prime\in\mathbb{Q})$ is true if there exists a collision-free path between configurations $q$ and $q^\prime$. Notice that the motion planning aspect is addressed through the evaluation of predicates containing typed variables, while task planning deals with discrete variables representing objects. 
\item $\mathcal{I}$ denotes a set of initial literals. For ease of presentation, we also refer to a tuple of literals and the state of typed variables as a state. Therefore, $\mathcal{I}$ along with assigned values of typed variables describe the initial state of the world.
\item $\mathcal{G}$ denotes a conjunctive set of goal literals. 
\item $\Delta$ denotes a finite set of \emph{actions}, whose arguments are tuples of object variables and typed variables. Each action $\delta\in\Delta$ is described by a set of literal \emph{preconditions} $\textsl{pre}(\delta)$ and a set of literal \emph{effects} $\textsl{eff}(\delta)$. We use $+$ and $-$ as superscripts on \textsl{pre} and \textsl{eff} to represent positive and negative literals, respectively. An action with an assignment of values is applicable to a state if $\textsl{pre}^+(\delta)\subseteq\mathcal{I}$ and $\textsl{pre}^-(\delta)\cap\mathcal{I}=\emptyset$. The successor state becomes a tuple of $\mathcal{I}\setminus\textsl{eff}^-(\delta)\cup\textsl{eff}^+(\delta)$ along with the values of typed variables assigned in the action.
\end{itemize}

A solution to the TAMP problem is a finite sequence of actions from the initial state, specified by the initial literals $\mathcal{I}$, to reach the goal state, satisfying the goal literals $\mathcal{G}$. A partial plan, where object variables in a sequence of actions in the plan are only determined (often called \emph{grounding}) while continuous typed variables are not yet chosen (often called \emph{refinement}), is referred to as a \emph{plan skeleton}~\cite{lozano2014constraint}. Predicates involving only typed variables, such as $\texttt{Reachable}(q\in\mathbb{Q}, q^\prime\in\mathbb{Q})$, are always considered true when finding plan skeletons. 

Refinement of actions in a plan skeleton, involving the evaluation of $\texttt{Reachable}(q\in\mathbb{Q}, q^\prime\in\mathbb{Q})$ predicates, finds valid robot motions that satisfy geometric and kinematic constraints, such as arm motion for grasping a cup and base motion to reach a table. Geometric constraints ensure that the robot does not collide with any objects in the world or itself at any time, and that no collision occurs among any pairs of objects. Kinematic constraints govern the degrees of freedom of the robot and the relationships among links connected through joints. 

A finite sequence of valid motions obtained by refinement constitutes a continuous path in $\mathbb{Q}$ that the robot follows from its initial state to reach a goal. Given a specification of the robot's motion model, execution time can be computed from the resultant path. Optimal planners aim at finding a plan that minimizes the total execution time~\cite{vega2020asymptotically,shome2021pushing,thomason2022task}. However, in this work, our goal is to identify any plan that can be executed before a deadline, while also taking into account the planning (or computation) time required to identify it.

In Section~\ref{sec:example}, we provide example scenarios in navigation and manipulation domains to demonstrate TAMP problem specifications in practice.

\subsection{Effort allocation for deadline-aware TAMP problems}
\label{subsec:meta}

\begin{figure*}[htbp]
    \centering
    \begin{minipage}{0.7\textwidth}
        \centering
        \begin{subfigure}[b]{0.45\textwidth}
            \centering
            \includegraphics[width=\textwidth]{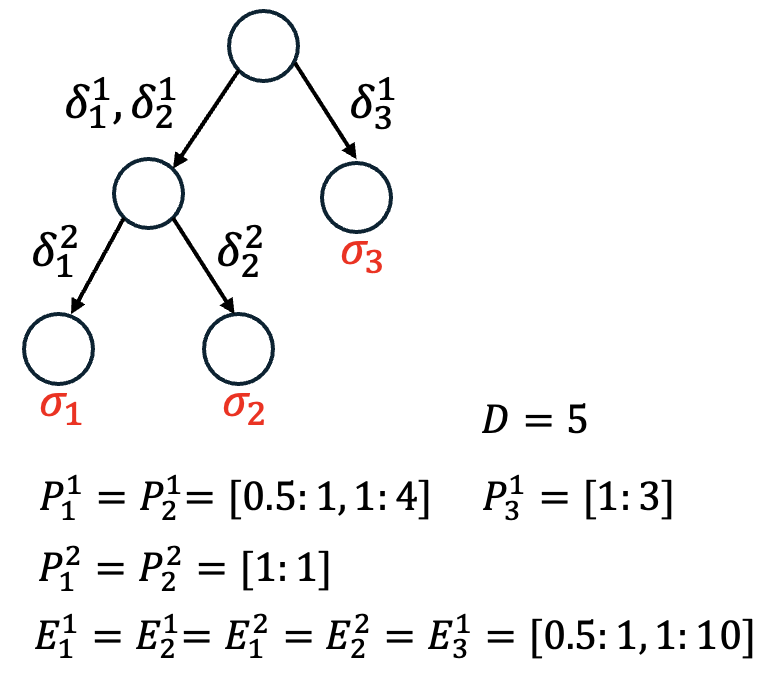}
            \caption{Problem instance for this example. Here, probabilities are represented in CDF form, where [CDF probability: time steps required for planning or execution] is interpreted as the probability that the planning or execution takes less than or equal to the specified time step.
            }
            \label{fig:instance}
        \end{subfigure}
        \hfill
        \begin{subfigure}[b]{0.45\textwidth}
            \centering
            \includegraphics[width=\textwidth]{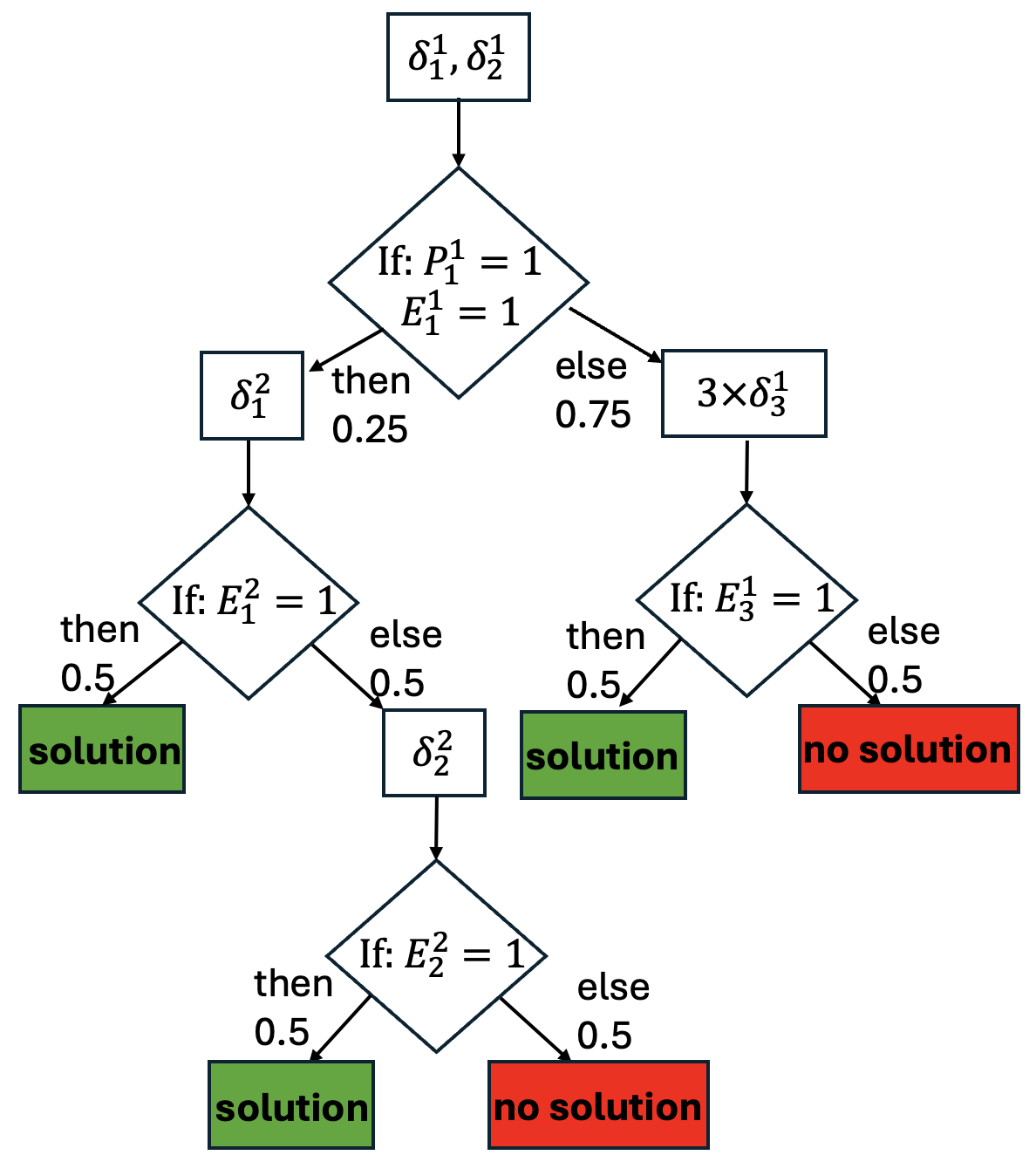}
            \caption{Optimal policy found for this example. With a slight abuse of notation, values of $P$ and $E$ denote time steps. Values on the edges denote transition probabilities. $3\times \delta_3^1$ implies selecting $\delta_3^1$ for three time steps.
            }
            \label{fig:optimal_policy}
        \end{subfigure}
    \end{minipage}
    
    \vspace{0.5cm} 

    \begin{subfigure}[b]{0.67\textwidth}
        \centering
        \includegraphics[width=\textwidth]{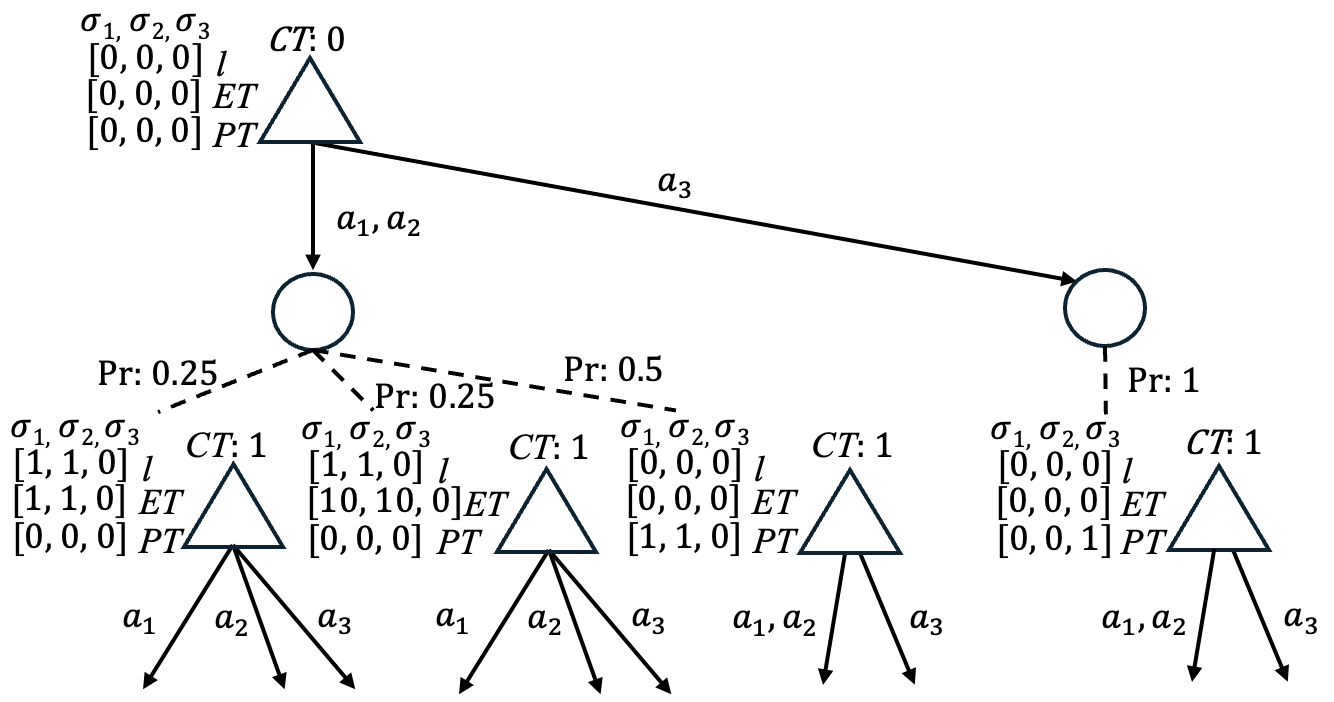}
        \caption{A partial representation of the MDP by an expectimax tree generated for this example.}
        \label{fig:tree}
    \end{subfigure}
    
    \caption{Example of an effort allocation problem.}
    \label{fig:example}
\end{figure*}

The TAMP problems introduced above do not take into account the notion of a deadline or time budget, so their solvers only aim at finding a satisfying plan regardless of the time it takes to find it. However, in this work, we explicitly address deadline-aware planning, which imposes an additional constraint ensuring that a plan found must be refinable and executable within a deadline.  

Our work adheres to the \emph{sequence-before-satisfy} approach~\cite{wolfe2010combined,srivastava2014combined,toussaint2015logic,lagriffoul2016combining,dantam2018incremental,lo2020petlon,garrett2020pddlstream,sung2023learning}, often leveraging AI heuristics, such as Fast Downward~\cite{helmert2006fast}, to efficiently find plan skeletons initially without considering continuous typed variables. The search for plan skeletons is then followed by finding satisfying values for the typed variables. 

Since plan skeleton search is generally computationally tractable compared to continuous value selection, one can first find $K$ plan skeletons to jointly search for continuous values satisfying constraints, a process referred to as top-$K$ planning~\cite{speck2020symbolic,ren2021extended}. This approach may lead to finding a solution quickly, as any value selection in some of the plan skeletons may violate the constraints, and thus, not yield solutions. Top-$K$ plan skeletons can be chosen by introducing either a user-defined cost or unit cost on every action. In the latter case, $K$ plan skeletons are selected starting from the smallest number of actions reaching the goal in ascending order.

In this work, we consider a scenario where multiple plan skeletons are provided, but the planning (or refining) time and execution time of individual actions within the plan skeletons are \emph{unknown}. It is important to note that we interchangeably use (motion) ``planning" and ``elaboration" to mean refinement throughout the paper. Let $\Sigma=\{\sigma_1, ..., \sigma_k, ..., \sigma_K\}$ be a finite set of top-$K$ plan skeletons. Each plan skeleton $\sigma_k$ contains $A_k$ actions, denoted by $\sigma_k=(\delta_k^1, ..., \delta_k^j,..., \delta_k^{A_k})$, where $\delta_k^j$ denotes the $j$-th action in plan skeleton $\sigma_k$.

We consider \emph{discrete} time steps for both planning time and execution time, where the interval between sequential time steps is determined by a user. Let $\{1, ..., D\}$ denote a set of decision-making time steps, where the planner decides which plan skeleton to allocate one interval of its time (or effort) to refine a corresponding action, that is, finding satisfying continuous values of typed variables of the action. Multiplying a user-defined time interval by $D$ yields the total wall-clock time allotted to the planner, corresponding to a \emph{deadline}. 

To handle unknown planning
time and execution time, we propose a learning paradigm for estimating these quantities using collected data obtained from past experiences. We round both the planning time and execution time of every action to match with decision-making time steps and introduce discrete distribution functions to model the uncertainty about these quantities. Specifically, we define $\mathrm{P}_k^j$, a discrete cumulative distribution function (CDF) modeling a \emph{planning time distribution} required to find valid continuous values of typed variables that satisfy corresponding constraints, and $\mathrm{E}_k^j$ a discrete CDF modeling an \emph{execution time distribution} required for the robot to execute a motion obtained by the refined action. For convenience, we define $\mathrm{p}_k^j$ and $\mathrm{e}_k^j$ as the probability mass functions (PMF) corresponding to $\mathrm{P}_k^j$ and $\mathrm{E}_k^j$, respectively. These PMFs can be computed directly from their corresponding CDFs.


Note that once the planner computes a motion, the execution time to execute the motion can be obtained deterministically as it depends on the path length. However, before a path is obtained, the execution time is stochastic. The stochasticity arises from the random process of generating motions in motion planning.  

We finally propose an \emph{effort allocation problem}, whose objective is to maximize the probability of finding a fully refined plan skeleton that can be planned and executed by the robot within a pre-specified deadline, under uncertain planning times and execution times, by learning a computation allocation policy among plan skeletons.

An example of effort allocation for a deadline-aware TAMP problem instance is illustrated in Figure~\ref{fig:example} (1). In this example, there are three plan skeletons: $\sigma_1 = (\delta_1^1, \delta_1^2)$, $\sigma_2 = (\delta_2^1, \delta_2^2)$, and $\sigma_3 = (\delta_3^1)$. Note that $\sigma_1$ and $\sigma_2$ share the first action, \ie, $\delta_1^1=\delta_2^1$. The deadline $D$ is 5 time steps. All actions have the same execution time probability mass function: they finish in 1 time step 50\% of the time and in 10 time steps the other 50\% of the time. Additionally, in this example, actions $\delta_1^2$, $\delta_2^2$, and $\delta_3^1$ have deterministic planning durations: three steps for $\delta_3^1$ and one step each for $\delta_1^2$ and $\delta_2^2$. Action $\delta_1^1$ (or $\delta_2^1$) has a stochastic planning duration, taking either one or four time steps with equal probability.

The optimal policy in this scenario, depicted in Figure~\ref{fig:example} (2) is to first execute the motion planner on $\delta_1^1$ (or $\delta_2^1$) for 1 time step. If motion planning has not found a plan, or if it found a plan requiring 10 time steps for execution, the motion planner then proceeds to refine $\delta_3^1$ for three time steps. This refinement is guaranteed to succeed; however, the resulting plan can be executed on time only if the execution time is 1 step (not 10); otherwise, no solution is found. If $\delta_1^1$ (or $\delta_2^1$) completes planning within 1 time step and discovers a motion plan executable in 1 time step, the motion planner continues to refine $\delta_1^2$ for 1 time step. If this plan can be executed in 1 time step, a solution is found. Otherwise, the motion planner proceeds to refine $\delta_2^2$ for 1 time step. If the resulting plan can be executed in 1 time step, a solution is found; otherwise, no solution is found.
The overall success probability of this policy is $0.5625$. This value can be extracted from the figure by multiplying the probabilities along each path leading to a leaf node where a solution is found and then summing the probabilities across these different paths.


In the following subsection, we finalize our problem formulation using the notation introduced so far.

\subsection{An MDP model}
\label{subsec:mdp}

We formalize the effort allocation problem using a Markov decision process (MDP) model, which can be represented by a tuple $\langle\mathcal{S}, \mathcal{A}, \mathrm{T}, \mathrm{R}, D\rangle$ as follows:
\begin{itemize}
\item $\mathcal{S}$ denotes a finite set of MDP states, where $\mathcal{S}=\{(CT, l_1,$ $ ..., l_k, ..., l_K, PT_1, ..., PT_k,$ $ ..., PT_K, ET_1, ..., ET_k, ...,$ $ ET_K), \texttt{success}, \texttt{failure}\}$. Here, $CT\in\mathbb{N}$ denotes the current time step. $l_k\in\{0, ..., A_k\}$ is an index of the last action in plan skeleton $\sigma_k$ that has already been refined to find a path, where $l_k = 0$ indicates that no actions in the plan skeleton have been refined yet. $PT_k\in\mathbb{N}$ denotes accumulated time steps spent for planning so far for action $\delta_k^{l_k+1}$ in plan skeleton $\sigma_k$. $PT_k$ maintains accumulated time steps for action $\delta_k^{l_k+1}$ specifically instead of action $\delta_k^{l_k}$ because the planner only needs to keep track of planning time for \emph{unrefined} actions. For actions that are refined, the planner has already confirmed the actual planning times, so refined actions no longer have stochasticity. $ET_k\in\mathbb{N}$ denotes accumulated time steps that will be spent for execution for all actions refined so far in plan skeleton $\sigma_k$. \texttt{success} and \texttt{failure} denote the terminal states, where \texttt{success} implies the successful refinement of one of the $K$ plan skeletons before the deadline, and \texttt{failure} implies the opposite.
\item $\mathcal{A}=\{a_1, ..., a_k, ..., a_K\}$ denotes a finite set of MDP actions, where MDP action $a_k$  selects plan skeleton $\sigma_k$ to spend one time step refining action $\delta_k^{l_k+1}$.

\item $\mathrm{T}$ is a transition function, where $\mathrm{T}(s\in\mathcal{S}, a_k\in\mathcal{A}, s^\prime\in\mathcal{S})$ represents the probability of transitioning from MDP state $s=(CT, l_1,$ $ ..., l_k, ..., l_K, PT_1, ..., PT_k, ..., PT_K, ET_1,$ $ ..., ET_k, ..., ET_K)$ to MDP successor state $s^\prime=(CT^\prime, l_1^\prime, ..., l_k^\prime, ..., l_K^\prime, PT_1^\prime, ..., PT_k^\prime, ..., PT_K^\prime, $ $ET_1^\prime, ...,$ $ ET_k^\prime, ..., ET_K^\prime)$ by taking MDP action $a_k$. We define four types of transitions as follows: 
\begin{enumerate}
\item $CT^\prime = CT + 1$ with a probability of $1$. We add an additional time step for each decision-making time step. 
\item $l_m^\prime, PT_m^\prime, ET_m^\prime=l_m, PT_m, ET_m$ with a probability of $1$ for all $m$ such that $\delta_m^{l_m+1}\neq\delta_k^{l_k+1}$. 
\item For action $\delta_k^{l_k+1}$, a path can be found with a probability of $\mathrm{P}_k^{l_k+1}(PT_k+1)$. For all $m$ such that $\delta_m^{l_m+1}=\delta_k^{l_k+1}$, if $l_m<A_k$, then $l_m^\prime, PT_m^\prime, ET_m^\prime=l_m+1, 0, ET_m+x$, where $x\in$ $\texttt{support}(\mathrm{e}_k^{l_k+1})$\footnote{The \texttt{support} refers to the subset of the domain of a probability mass function $\mathrm{E}$ where the probability measure is non-zero.}, with a probability of $\mathrm{P}_k^{l_k+1}(PT_k+1)\cdot\mathrm{e}_k^{l_k+1}(x)$, or \texttt{success} otherwise.
\item Alternatively, a path cannot be found with a probability of $1-\mathrm{P}_k^{l_k+1}(PT_k+1)$ even after spending the current time step. In such a case, for all $m$ such that $\delta_m^{l_m+1}=\delta_k^{l_k+1}$, if $CT<D$, then $l_m^\prime, PT_m^\prime, ET_m^\prime=l_m, PT_m+1, ET_m$, or \texttt{failure} otherwise.
\end{enumerate}

\item $\mathrm{R}$ is a reward function. The reward is $1$ if the transition reaches the \texttt{success} terminal state, or $0$ if it reaches the \texttt{failure} terminal state.
\end{itemize}

The objective of this MDP formulation is to find an optimal policy that maps each state from $\mathcal{S}$ to an action from $\mathcal{A}$, leading to maximizing the expected cumulative reward.

It is important to note that our formulation does not require assumptions of \emph{independent} plan skeletons or \emph{downward refinability}~\cite{marthi2007angelic}. Even if an MDP action chooses plan skeleton $\sigma_k$ to refine a corresponding action $\delta_k^{l_k+1}$, for all $m$ such that $m\ne k$ but $\delta_m^{l_m+1}=\delta_k^{l_k+1}$, the MDP updates the state for plan skeleton $\sigma_m$ as well, thereby considering dependent plan skeletons. Downward refinability implies that plan skeletons are always refinable. However, our formulation can handle cases where a planning time distribution $\mathrm{P}_k^j$ does not always reach $1$ even at the deadline, meaning that not all actions may be refinable. This capability effectively removes the downward refinability assumption and can address practical scenarios.

An example of the MDP is depicted in Figure~\ref{fig:example} (3). The MDP is represented as an expectimax tree, where the initial state, with no time allocated to any plan skeleton, is the root. Only the first two levels of the tree are presented. From the root, the available actions are to either attempt to refine $\delta_1^1$ or equivalently $\delta_2^1$ by an MDP action $a_1$ or equivalently $a_2$, or to try and refine $\delta_3^1$ by an MDP action $a_3$.
If 1 time step is invested in refining $\delta_3^1$ by an MDP action $a_3$, the motion planner cannot find a plan, resulting in only one successor state where the planning time for $\sigma_3$ is increased by 1, with all other values remaining unchanged.
If 1 time step is invested in refining $\delta_1^1$ or equivalently $\delta_2^1$ by an MDP action $a_1$ or equivalently $a_2$, the motion planner does not finish planning with a probability of 0.5. In this case, the planning time for $\sigma_1$ and $\sigma_2$ is increased by 1 in the resulting state, with all other values unchanged. With a probability of 0.5, the refinement of action $\delta_1^1$ or equivalently $\delta_2^1$ is completed. Since the execution time of the refined plan can either be 1 or 10 (with equal probabilities), there are two possible resulting states. Both states will have $l_1 = 1$, $PT_1 = 0$, $l_2 = 1$, and $PT_2 = 0$, indicating that $\delta_1^1$ or equivalently $\delta_2^1$ has finished refinement. The first state has $ET_1 = ET_2 = 1$, while the second state has $ET_1 = ET_2 = 10$.

We also note that generalization of the proposed problem, making one plan skeleton more favorable than another even if both meet a deadline, is achievable by designing more complex MDP specifications. For instance, one could attach literals to MDP states and assign a reward to specific literals, such as reaching a meeting room before a deadline with a coffee grabbed, making that occurrence more favorable than merely making a meeting before the deadline.

The effort allocation problem reveals \emph{metareasoning} perspectives as it allows efficient use of limited computational resources (\ie, meta-level computation) for finding a solution to the TAMP problems (\ie, object-level computation). Essentially, the effort allocation problem addresses when and what to consider, when to stop deliberation, and how to adaptively allocate computation under a limited computational budget (\ie, deadline).

\section{Theoretical Analysis}
\label{sec:hardness}
As shown in the previous section, the effort allocation problem can be formulated as an MDP. MDPs can be efficiently solved, for instance, using methods like value iteration, which have polynomial-time complexity with respect to the number of states and actions. However, the challenge arises from the number of states in the MDP corresponding to all conceivable time allocations and their associated outcomes, which grows exponentially with the size of the input problem.

In this section, we establish that the effort allocation problem is NP-hard, even when the execution time is known. This finding suggests that addressing this problem necessitates the use of approximation algorithms or heuristic approaches to yield feasible solutions within reasonable computational bounds.

\begin{theorem} Finding the optimal policy for the effort allocation problem is NP-hard.
\label{thm:np_hard}
\end{theorem}
\noindent \textbf{Proof:} We establish NP-hardness by reducing from the optimization version of the knapsack problem.

\begin{definition}[Knapsack problem{~\cite[problem MP9]{GareyJohnsson}}]
Given a finite set of items $\mathcal{S}=\{s_1, ..., $ $s_k, ..., s_K\}$, each with a positive integer weight $w_k$ and value $v_k$, and a weight limit $W$, find a subset $\mathcal{S}^\circ$ of $\mathcal{S}$ such that the total weight of $\mathcal{S}^\circ$ is at most $W$ and the total value of $\mathcal{S}^\circ$ is maximized.
\end{definition}

Without loss of generality, we assume that all the values are greater than or equal to 1. That is, $v_k \geq 1$ for all $1 \leq k \leq K$. We also assume, without loss of generality, that $\mathcal{S}^\circ=\{s_1, ..., s_m, ..., s_M\}$ with $M\le K$.

In our reduction, each plan skeleton represents an item in the knapsack problem. We only consider degenerate plan skeletons $\Sigma=\{\sigma_1, ..., \sigma_k, ..., \sigma_K\}$, where each plan skeleton includes only a single action, and actions do not overlap between plan skeletons. Formally, for all $1 \leq k \leq K$, $\sigma_k = (\delta_k^1)$ and $\delta_k^1 \neq \delta_m^1$ for all $k \neq m$. Additionally, we assign a zero execution time for each plan skeleton, that is, $\mathrm{E}_k^1 = 0$ with probability 1, and set a deadline $D=W$.\footnote{The proof can be easily adapted to support any constant execution time.} Finally, the planning time CDF $\mathrm{P}_k^1$ is defined as a piecewise constant function:
 		\begin{align*}
 		&\mathrm{P}_k^1(t)  = 
 		& \left\{\begin{array}{lr}
 		0, & t < w_k,\\
 		\epsilon v_k, & w_k \leq t \leq W, \\
        1,                & W < t.\\
		\end{array}\right.
		\end{align*}
  
Here, plan skeleton $k$ has a probability of 0 of completing elaboration and execution before time $w_k$, a probability of $\epsilon v_k$ of completing elaboration and execution after time $w_k$, and the probability increases further to 1 after the deadline. Later in the proof, we set a particular value for $\epsilon$ such that the probability of success of at least one plan skeleton is \emph{almost} as much as the sum of the success probabilities of all the selected skeletons.
 
To prove NP-hardness, we establish bounds on the probability of success, denoted $P_{SUCC}(\mathcal{S}^\circ)$, for any set of items $\mathcal{S}^\circ$:  \begin{align*}\label{eq:prob_bounds}
 \epsilon \sum_{s_m\in \mathcal{S}^\circ}  v_m &\geq
 P_{SUCC}(\mathcal{S}^\circ), \\
 &= 1-\prod_{s_m\in \mathcal{S}^\circ} (1-\epsilon v_m), \\ 
 &> \epsilon ((\sum_{s_m\in \mathcal{S}^\circ} v_m)-1).
 \end{align*}

The first inequality follows from the union bound, stating that the sum of event probabilities, where the probability of each event for plan skeleton $m$ corresponds to $\epsilon v_m$, is always at least as large as the probability of the union of these events, implying that at least one plan skeleton finishes elaboration and execution. 

The last inequality is demonstrated by bounding the higher-order terms when expanding the product into a sum. Now, let's proceed with the detailed proof for this inequality. 
First, we can derive the following equality using the principle of inclusion-exclusion:
\begin{align*}
P_{SUCC}(\mathcal{S}^\circ)&=1-\prod_{m=1}^M (1-\epsilon v_m), \\
&= \sum_{m=1}^M \epsilon^m(-1)^{m+1}
 \sum_{\substack{\mathcal{N} \subseteq \{1, ..., M\}, \\ |\mathcal{N}|=m}}
   \prod_{n\in \mathcal{N}} v_n.
\end{align*}

All the summands in the outer summation with odd $m$ are positive, so if we exclude them (for all $m>1$), the value of the expression does not increase.

\begin{align*}
&P_{SUCC}(\mathcal{S}^\circ)= \sum_{m=1}^M \epsilon^m(-1)^{m+1}
 \sum_{\substack{\mathcal{N} \subseteq \{1, ..., M\}, \\ |\mathcal{N}|=m}}
   \prod_{n\in \mathcal{N}} v_n, \\
   &\geq \epsilon
 \sum_{\substack{\mathcal{N} \subseteq \{1, ..., M\}, \\ |\mathcal{N}|=1}}
   \prod_{n\in \mathcal{N}} v_n - \sum_{m'=1}^{\lfloor \frac{M}{2} \rfloor} \epsilon^{2m'}
 \sum_{\substack{\mathcal{N} \subseteq \{1, ..., M\}, \\ |\mathcal{N}|=2m'}}
   \prod_{n\in \mathcal{N}} v_n, \\
   &= \epsilon \sum_{m=1}^M  v_m - \sum_{m'=1}^{\lfloor \frac{M}{2} \rfloor} \epsilon^{2m'}
 \sum_{\substack{\mathcal{N} \subseteq \{1, ..., M\}, \\ |\mathcal{N}|=2m'}}
   \prod_{n\in \mathcal{N}} v_n.
\end{align*}


On the other hand, the number of elements in $\sum_{\substack{\mathcal{N} \subseteq \{1, ..., M\}, \\ |\mathcal{N}|=2m'}}
   \prod_{n\in \mathcal{N}} v_n$ is $  {M}\choose{2m'}$, which is upper-bounded by $M^{2m'}$. Let $H=\max_{m=1}^M v_m$.
   Since $H \geq v_m$ for all $1 \leq m \leq M$, we have $H^{2m'} \geq \prod_{n\in \mathcal{N}} v_n$. We obtain the following inequalities using these relations:
\begin{align*}
P_{SUCC}(\mathcal{S}^\circ)&\geq \epsilon \sum_{m=1}^M  v_m - \sum_{m'=1}^{\lfloor \frac{M}{2} \rfloor} \epsilon^{2m'}
 \sum_{\substack{\mathcal{N} \subseteq \{1, ..., M\}, \\ |\mathcal{N}|=2m'}}
   \prod_{n\in \mathcal{N}} v_n, \\
   &\geq \epsilon \sum_{m=1}^M  v_m - \sum_{m'=1}^{\lfloor \frac{M}{2} \rfloor}{\epsilon ^{2m'} \sum_{\substack{\mathcal{N} \subseteq \{1, ..., M\}, \\ |\mathcal{N}|=2m'}} H^{2m'}}, \\
   &\geq \epsilon \sum_{m=1}^M  v_m - \sum_{m'=1}^{\lfloor \frac{M}{2} \rfloor}{(\epsilon H M) ^{2m'}}.\\
\end{align*}


By setting $\epsilon = \frac{1}{H^2M^3}$, we obtain:
\begin{align*}
P_{SUCC}(\mathcal{S}^\circ)
&\geq \epsilon \sum_{m=1}^M  v_m - \sum_{m'=1}^{\lfloor \frac{M}{2} \rfloor}{(\epsilon H M) ^{2m'}}, \\
&= \epsilon \sum_{m=1}^M  v_m - \epsilon \sum_{m'=1}^{\lfloor \frac{M}{2} \rfloor}{\epsilon^{2m'-1}(HM)^{2m'}}, \\
&\geq \epsilon (\sum_{m=1}^M  v_m -\sum_{m'=1}^{\lfloor \frac{M}{2} \rfloor}{\frac{(HM)^{2m'}}{(H^2M^3)^{2m'-1}}}), \\
&= \epsilon (\sum_{m=1}^M  v_m -\sum_{m'=1}^{\lfloor \frac{M}{2} \rfloor}{\frac{1}{H^{2m'-2}M^{4m'-3}}}), \\
&\geq \epsilon (\sum_{m=1}^M  v_m -\sum_{m'=1}^{\lfloor \frac{M}{2} \rfloor}{\frac{1}{M}}), \\
&> \epsilon ((\sum_{s_m\in \mathcal{S}^\circ} v_m)-1),
\end{align*}
where the second-to-last inequality can be derived from the fact that $\frac{1}{H^{2m'-2}M^{4m'-3}} \le \frac{1}{M}$, since $H \ge v_k \ge 1$, $M \ge 1$, and $m' \ge 1$.

Next, we demonstrate that optimal schedules correspond to optimal knapsack solutions. Let $\mathcal{S}^\circ$ be an optimal schedule. We first show that $\mathcal{S}^\circ$ corresponds to a knapsack solution. Since the deadline of all plan skeletons is $W$, the processing time assigned to each plan skeleton $s_m\in \mathcal{S}^\circ$ cannot exceed $W$. Therefore, any plan skeleton assigned non-zero processing time in $\mathcal{S}^\circ$ receives time equal to $w_m$. Additionally, the sum of processing times in $\mathcal{S}^\circ$ is at most $W$. We use $\mathcal{S}^\circ$ to also denote the set of items in the knapsack problem corresponding to the plan skeletons, each assigned time $w_m$ in $\mathcal{S}^\circ$. The knapsack value of $\mathcal{S}^\circ$ is $V=\sum_{s_m\in \mathcal{S}^\circ} v_m$, and since the sum of weights of the items in $\mathcal{S}^\circ$ is at most $W$, $\mathcal{S}^\circ$ is a knapsack solution.

The remaining proof is to show that $\mathcal{S}^\circ$ is an optimal knapsack solution. Assume, in contradiction, that $\mathcal{S}^\circ$ is suboptimal as a knapsack solution. Then there must exist a knapsack solution $\mathcal{S}^+$ with value $V^+ \geq V+1$. Since $\mathcal{S}^+$ is a knapsack solution, $\sum_{s_m\in \mathcal{S}^+} w_m \leq W$. Thus, taking $\mathcal{S}^+$ as a schedule (assigning time $w_m$ to each plan skeleton $s_m\in \mathcal{S}^+$) creates a schedule where all plan skeletons run before time $W$ as well, with success probability:
\begin{align*}
P_{SUCC}(\mathcal{S}^+) &= 1-\prod_{s_m\in \mathcal{S}^+} (1-\epsilon v_m)  > 
\epsilon ((\sum_{s_m\in \mathcal{S}^+} v_m)-1), \\
& \geq \epsilon (\sum_{s_m\in \mathcal{S}^\circ} v_i) \geq P_{SUCC}(\mathcal{S}^\circ),
\end{align*}
where the inequalities are due to the previously established bounds. That is, schedule $\mathcal{S}^+$ has a greater success probability than $\mathcal{S}^\circ$, a contradiction. $\Box$

\section{Model-Based Approach}
\label{sec:model}
Solving the proposed MDP-based effort allocation formulation requires access to planning time distributions and execution time distributions of all abstract actions involved in plan skeletons. Since these distributions are unknown a priori, we must either learn them from past experiences or data, or find a way to bypass explicit learning for finding a policy. 

To address this challenge, we explore both \emph{model-based} and \emph{model-free} approaches. In model-based learning, we collect data on both planning times and execution times of all abstract actions taken in the training problems and estimate their distributions. In model-free learning, we utilize policy optimization from reinforcement learning to bypass distribution learning. We will delve into the model-free approach in the next section. 

Both planning time distribution $P_k^j$ and execution time distribution $E_k^j$ are discrete, each modeled as a \emph{categorical} or \emph{multinoulli} distribution with parameter vector $\theta_{k,j}^{P/E}$ in the $D$-simplex, as each distribution contains $D+1$ categories as a support. $D$ parameters are positive real that sum to $1$. The value of $D+1$ implies non-plannable or non-executable within a deadline, relaxing the downward refinability assumption explained in Section~\ref{subsec:mdp}.

We compute maximum likelihood estimation of $\theta_{k,j}^{P/E}$ parameters for all $k$ and $j$ by counting the number of the corresponding events in the data. Laplace smoothing can also be applied to make robust estimation if data is sparse.

Using the estimated distributions, we are now prepared to solve the proposed MDP problem. This problem-solving incurs meta-level computation, where smaller is better as the planner can allocate its resources more efficiently to object-level computations, corresponding to planning and execution times. However, as demonstrated in Section~\ref{sec:hardness}, optimally solving the proposed MDP using, for example, value iteration, is computationally intractable. Therefore, we propose two approximate schemes for efficiently solving the MDP.

The first approach is approximating value functions via MCTS. The second approach is to compute the optimal \emph{linear contiguous} policy, which is a linear policy in which all time allocations to the same plan skeleton are performed contiguously.
\subsection{Monte Carlo Tree Search}
We employ MCTS~\cite{browne2012survey,swiechowski2023monte} to handle the exponential increase of states in the proposed MDP, as shown in Theorem~\ref{thm:np_hard}. To overcome the curse of dimensionality, MCTS leverages random episode sampling within a decision tree to approximate an optimal policy that would be found by an expensive value iteration algorithm.

A decision tree of MCTS when applied to solving MDP problems corresponds to an \emph{expectimax} tree, where the tree nodes, starting from the root, are structured as an alternating sequence of action-choice nodes and probabilistic nodes. The value of a probabilistic node is computed as the expectation of the values of its children nodes, and the value of an action-choice node is computed as the maximum of the values of its children. The unique feature of the expectimax tree constructed for the proposed MDP is that every action-choice node always has $K$ children nodes, as the action is about determining which plan skeleton to refine among $K$ plan skeletons regardless of the state being considered, except for terminal states.

Value iteration utilizes the optimal state-action value function $Q^*:\mathcal{S}\times \mathcal{A}\rightarrow\mathbb{R}$, defined as:
\begin{equation*}\label{eqn:q_funct}
Q^*(s, a) = R(s, a) + \sum_{s^\prime\in\mathcal{S}} \mathrm{T}(s, a, s^\prime)V^*(s^\prime),
\end{equation*}
where $V^*$ denotes the optimal state value function. An optimal policy can then be defined as $\pi(s)=\argmax_{a\in\mathcal{A}} Q(s, a)$.

On the other hand, MCTS approximates the state-action value function $Q$ using random simulation. In particular, we adopt the upper confidence bounds applied for trees (UCT~\cite{kocsis2006bandit}) as an MCTS algorithm, which treats action choice as a $K$\emph{-armed bandit} problem for choosing from $K$ plan skeletons and aims to achieve the property that the probability of selecting a sub-optimal action converges to $0$ as the computation time goes to infinity by balancing exploration-exploitation trade-off. This trade-off is captured in the form of:
\begin{equation*}\label{eqn:uct}
\pi(s)=\argmax_{a\in\mathcal{A}} \Big\{Q(s, a)+C\sqrt{\frac{\ln{N(s)}}{N(s,a)}}\Big\},
\end{equation*}
where $N(s)$ denotes the number of times state $s$ has been visited, and $N(s, a)$ denotes the number of times action $a$ has been sampled in state $s$, both in previous iterations. $C>0$ denotes an exploration constant, where an increased value implies encouraging exploration, while a decreased value implies encouraging exploitation.

We apply a general MCTS framework iteratively running four phases: (1) \emph{selection}, where a node in the tree that is not yet explored is selected based on the above UCT criteria, (2) \emph{expansion}, where a chosen node is expanded by applying an available action, (3) \emph{simulation}, where random actions are applied to the node expanded from the chosen node until it reaches a terminal state, and (4) \emph{backpropagation}, where the reward accumulated in the episode generated by the simulation phase is backpropagated up to the root node. The accumulated reward corresponds to the number of successful rollouts that meet a pre-specified deadline. The policy quality can primarily be determined by the allotted MCTS computation time and the number of rollouts used in the simulation phase. 

MCTS is an \emph{anytime} algorithm, implying that the more computation time is exploited, the closer the computed policy is to the optimal, and that it can be terminated at any time but still outputs an answer. This property is desirable as users can determine the meta-level computation time for their applications at the sacrifice of policy quality. It is worth noting that finding a sweet spot between MCTS computation time and policy quality can be considered as meta-meta-reasoning, but we do not delve into this aspect in this work.

\subsection{Heuristics}
\label{subsec:heuristics}
We introduce two properties that a policy may exhibit as heuristics, which systematically restrict the solution policy space. This restriction may result in identifying a different optimal policy than the one found by solving the proposed MDP, thereby forming an approximate solution, but it significantly facilitates the design of an efficient algorithm. The first property is \emph{linear}, and the second property is \emph{contiguous}.

A linear policy is a pre-determined allocation of time steps to all plan skeletons, which remains unchanged regardless of the refinement process unless a valid solution is found. The optimal policy in Figure~\ref{fig:example} (2) is not linear, as the decision of which action to allocate time to depends on the result of the previous allocation (\eg, we cannot know in advance whether action $\delta_1^2$ or $\delta_3^1$ will be allocated computation time after $\delta_1^1$ or $\delta_2^1$).

A contiguous policy is one in which all time allocations to the refinement of an action are performed contiguously, without any other allocations in between. The optimal policy in Figure~\ref{fig:example} (2) is contiguous, whereas the linear policy $(\delta_3^1, \delta_1^1, \delta_3^1)$, for example, is not contiguous.

Consider the special case where each plan skeleton consists of only a single action and the execution times are known. In this case, an action $\delta_k^j$ should never be executed if its execution time plus the current time exceeds the deadline, as it will never lead to a timely solution. Consequently, there exists an optimal solution that excludes such actions. When considering policies that only schedule computation actions if they can lead to a timely solution, the result of each computation allocation to an action $\delta_k^j$ is either: (1) a successful elaboration of action $\delta_k^j$, after which plan skeleton $k$ is fully elaborated and can be executed before the deadline, allowing the algorithm to terminate; or (2) action $\delta_k^j$ is not elaborated. Since only option (2) results in additional time allocations, the resulting policy is linear. Moreover, linear non-contiguous policies can be rearranged as contiguous without affecting the probability of success. Thus, in this special case, there exists an optimal policy that is both linear and contiguous.

While linear contiguous policies are not necessarily optimal in the general case, they can be computed in pseudo-polynomial time~\cite{shperberg2021situated}. Consequently, we can compute a linear contiguous policy and use it as a (potentially suboptimal) solution to the effort allocation problem to efficiently find a solution. 

To compute this policy, we introduce the dynamic programming (DP) scheme defined in Equation 1. $N(k, l_k+1)$ denotes the index set of plan skeletons that share the action $\delta_k^{l_k+1}$. For example, in Figure~\ref{fig:example} (1), $N(1, 1)=\{1, 2\}$ since $\delta_1^1=\delta_2^1$, and $N(3, 1)=\{3\}$. $PS$ represents a DP state, indicating the probability of successful refinement associated with that state. 

Specifically, $PS(k,0,0,0)$ represents the probability that plan skeleton $k$ will be completed on time under the linear contiguous policy. Equation~\ref{case1} serves as the termination condition, where the last action of the plan skeleton $\delta_k^{A_k}$ is being considered for refinement. All possible time allocations up to the deadline for $\delta_k^{A_k}$ are evaluated, with each allocation using the remaining time steps before the deadline as execution time, as no further refinement will be needed if it succeeds. Equation~\ref{case2} forms the core of the DP scheme. In this case, we again consider all possible time allocations up to the deadline for the current refining action $\delta_k^{l_k+1}$ (first summation). For each time allocation, we evaluate all possible execution times for $\delta_k^{l_k+1}$ (second summation). For each combination of time allocation and execution time, we apply the maximum value of the DP states that consider plan skeletons sharing the current action $\delta_k^{l_k+1}$, while adding the corresponding current and execution time steps. 

We compute the values of $PS(k, 0, 0, 0)$ for all plan skeletons and select the one with the highest value, indicating the most probable successful refinement. Once the skeleton is chosen by the DP, it is applied for actual refinement within the available time steps, ensuring the deadline is met.

\begin{figure*}
\begin{subnumcases}{PS(k, l_k, CT, ET_k)=}
    \sum\limits_{t=1}^{D-CT} p_k^{l_k+1}(t) \cdot E_k^{l_k+1}(D-t-CT-ET_k) & if $l_k+1 = A_k$, \label{case1}\\
    \begin{aligned}
    &\sum\limits_{t=1}^{D-CT} \sum\limits_{m \in \text{Support}(e_k^{l_k+1})} \Big(p_k^{l_k+1}(t)\cdot e_k^{l_k+1}(m)\\ 
    &\cdot \max_{k'\in N(k, l_k+1)} PS(k', l_{k'}+1, CT+t, ET_{k'}+m)\Big)
    \end{aligned}
    & if $l_k+1 < A_k$. \label{case2}
\end{subnumcases}

\end{figure*}

\begin{algorithm}[!ht]
\SetAlgoLined
\SetKwInOut{Input}{Input}
\SetKwInOut{Output}{Output}
\SetKw{Not}{not}
\SetKwFunction{SolveDP}{solveDP}
\SetKwFunction{Refine}{refine}
\SetKwFunction{UpdateDistribution}{updateDistribution}
\SetKwFunction{IsActionRefined}{isActionRefined}
\Input{$CT=0, l_1=0, ..., l_K=0, ET_1=0, ..., ET_K=0$}
\Output{\textsc{Success} or \textsc{Failure}}

\While{$CT + \min\{ET_1, ..., ET_K\} < D$}{

$\Sigma.\mbox{val}\leftarrow\emptyset$

\For{$0\le k \le K$}{
$\Sigma.\mbox{val}$.append$\Big($\SolveDP$\big(PS(k, l_k, CT, ET_k)\big)\Big)$ \tcp*[h]{Apply Equations~\ref{case1} and \ref{case2}.}
}

$k^\star\leftarrow \argmax_k{\Sigma.\mbox{val}}$

\Refine($\sigma_{k^\star}$) \tcp*[h]{Allocate one time step to refine the $k^\star$-th plan skeleton.}

\eIf{\IsActionRefined$(l_{k^\star} + 1)$} {
\If{$l_{k^\star} = A_{k^\star}$} {
\Return \textsc{Success}
}

$l_{k^\star}$++

$ET_{k^\star}\leftarrow ET_{k^\star}+m$ \tcp*[h]{Here, $m$ denotes the actual execution time derived from the refined action.}
}
{
\UpdateDistribution$(l_{k^\star} + 1)$ \tcp*[h]{Update the planning time distribution for the action $\delta_{{k^\star}}^{l_{k^\star} + 1}$.}
}

CT++
}

\Return \textsc{Failure}
\caption{DP\_Rerun}
\label{alg:dp_rerun}
\end{algorithm}

To mitigate the linearity assumption of DP, we introduce a variant of the algorithm called DP\_Rerun, with the pseudocode provided in Algorithm~\ref{alg:dp_rerun}. In this variant, at each decision-making time step, we first identify a plan skeleton that maximizes $PS(k, l_k, CT, ET_k)$ (lines 2-6), as done in DP. However, unlike DP, where the entire available time is allocated to the chosen skeleton, DP\_Rerun allocates only a single computational action for refinement (line 7). We then observe the result: either the elaboration of the corresponding action is completed, providing the execution time (lines 12, 13), or the elaboration process remains incomplete (line 15). In the latter case, we update the planning time distribution of the action, accounting for the time spent without completing the refinement. This process of running DP, selecting actions, obtaining observations, and updating the distribution continues iteratively until either the problem is solved (line 10) or the deadline has passed (line 19). Although DP\_Rerun requires more computation than DP due to its online replanning behavior, it mitigates the linear policy assumption and yields better solutions.

\section{Model-Free Approach}
\label{sec:free}

In the model-based approach introduced in the previous section, we estimate the parameters of planning time and execution time distributions using maximum likelihood estimation to learn the transition models $T$ in the MDP. This approach requires gathering data offline to fit the parameters of the distributions to match the data. The learned transition models are then used to compute a plan-skeleton selection policy for a query problem.

However, we often face situations where data is not readily available before the deployment of the robot, yet the robot still needs to learn a selection policy for a query problem without access to distribution information and improve its policy while repeatedly solving the problem. This approach essentially describes \emph{episodic reinforcement learning}. In this context, the agent operates on the same environment but is unaware of the underlying model (MDP). In this context, the agent interacts with the same environment but is unaware of the underlying model (MDP). The only information available to the agent is the state variables and the action space. The agent learns by observing the current state of the environment, performing an action, and observing the resulting next state and immediate reward.

In particular, we employ Proximal Policy Optimization (PPO~\cite{schulman2017proximal}), a prominent variant within the family of policy gradient methods, due to its desirable properties such as sample efficiency, stable training, and ease of use. PPO improves on traditional policy gradient methods by using a surrogate objective function that penalizes large policy updates, ensuring more stable and reliable learning. This is achieved through techniques such as clipping the probability ratio between the new and old policies, which prevents drastic changes that could destabilize training. Additionally, PPO uses mini-batch updates and adaptive step sizes, contributing to its effectiveness and efficiency in various reinforcement learning tasks.

In policy gradient methods, a policy is parameterized by a neural network that takes an MDP state as input and outputs an MDP action or a distribution of actions. In our context, the policy determines which plan skeleton to refine for one time step. The policy network is trained by deploying the agent in the environment for a finite collection of episodes. During these episodes, the agent collects sequences of states, actions, and rewards (trajectories), and the model is trained iteratively using this data.

It is worth mentioning, however, that the proposed reinforcement learning problem is intractable due to \emph{sparse} rewards. A reward of 1 is received only if the refinement of any plan skeleton is completed within a deadline, and a reward of 0 is received if it fails, as defined in the MDP described in Section~\ref{subsec:mdp}. Sparse reward problems, especially when the reward is only obtained at the end of the episode, are notoriously challenging. Without a reward signal until the episode's conclusion, it is difficult for the policy to learn which actions were beneficial, complicating the process of assigning credit and making effective decisions. Moreover, due to the stringent time deadlines inherent in these problems, exploration-promoting mechanisms like the Intrinsic Curiosity Module (ICM~\cite{PathakAED17}) and Go-Explore~\cite{goexplore}, which are typically effective in handling sparse rewards, are unlikely to be viable here. Each action taken for exploration decreases the probability of meeting the deadline, posing a significant challenge in balancing exploration against the imperative of timely completion.

\section{Example Scenarios}
\label{sec:example}

In this section, we provide two example scenarios implemented in PyBullet simulation~\cite{coumans2016pybullet} to demonstrate practical use cases of the proposed method and the specifications of the corresponding TAMP formulations. The first scenario is the navigation domain, where the goal is to reach a target office through a series of offices with different geometries. The second scenario is the manipulation domain, where the robot needs to move several objects to the kitchen area for meal preparation. In both scenarios, the planner must adhere to a pre-specified deadline. We will present experimental results using these domains in Section~\ref{subsec:eval_domains}.

\begin{figure*}[htbp]
\centering
\begin{subfigure}[b]{0.49\textwidth}
\centering
\includegraphics[width=\textwidth]{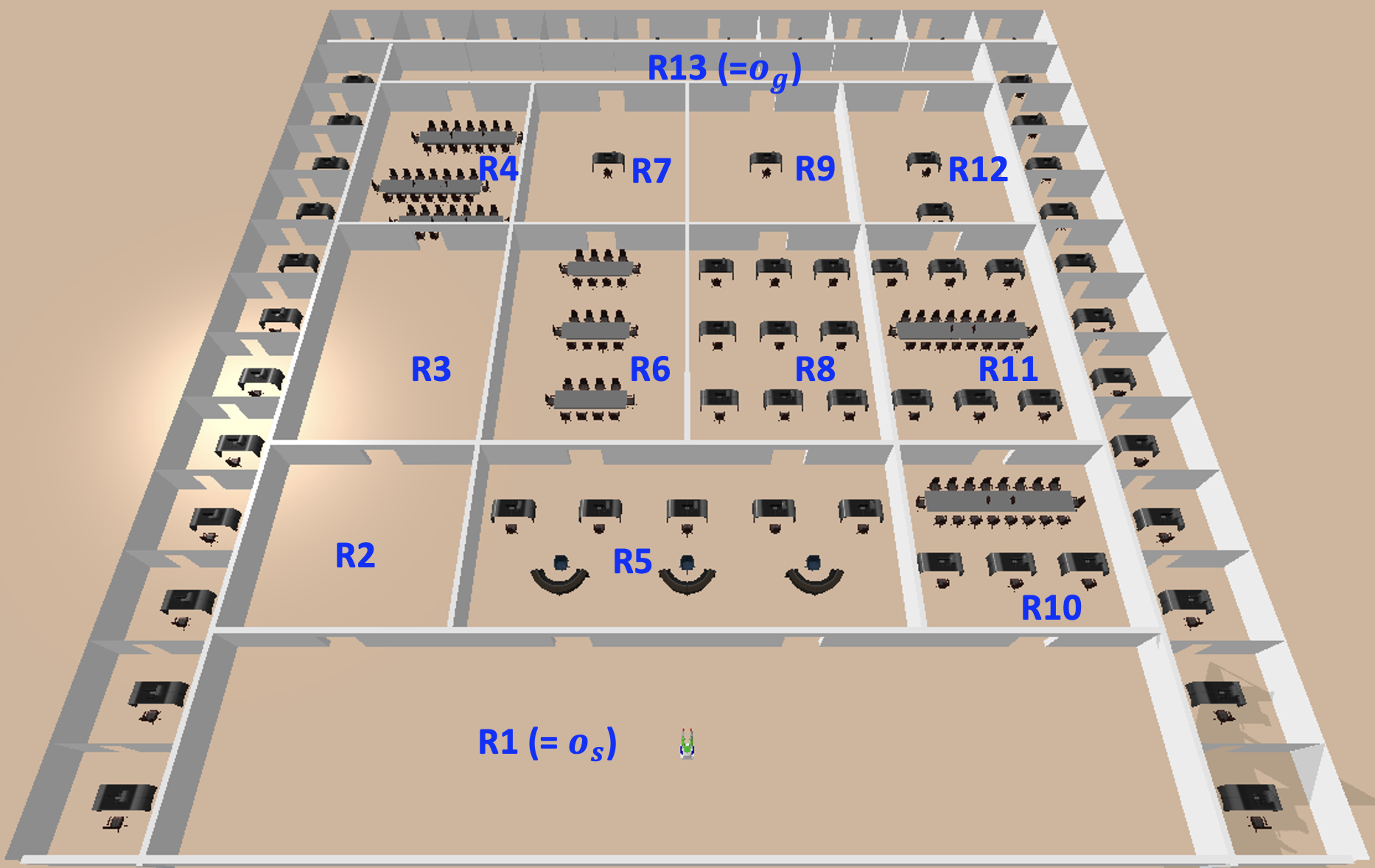}
\caption{Navigation domain.}
\end{subfigure}
\hfill
\begin{subfigure}[b]{0.50\textwidth}
\centering
\includegraphics[width=\textwidth]{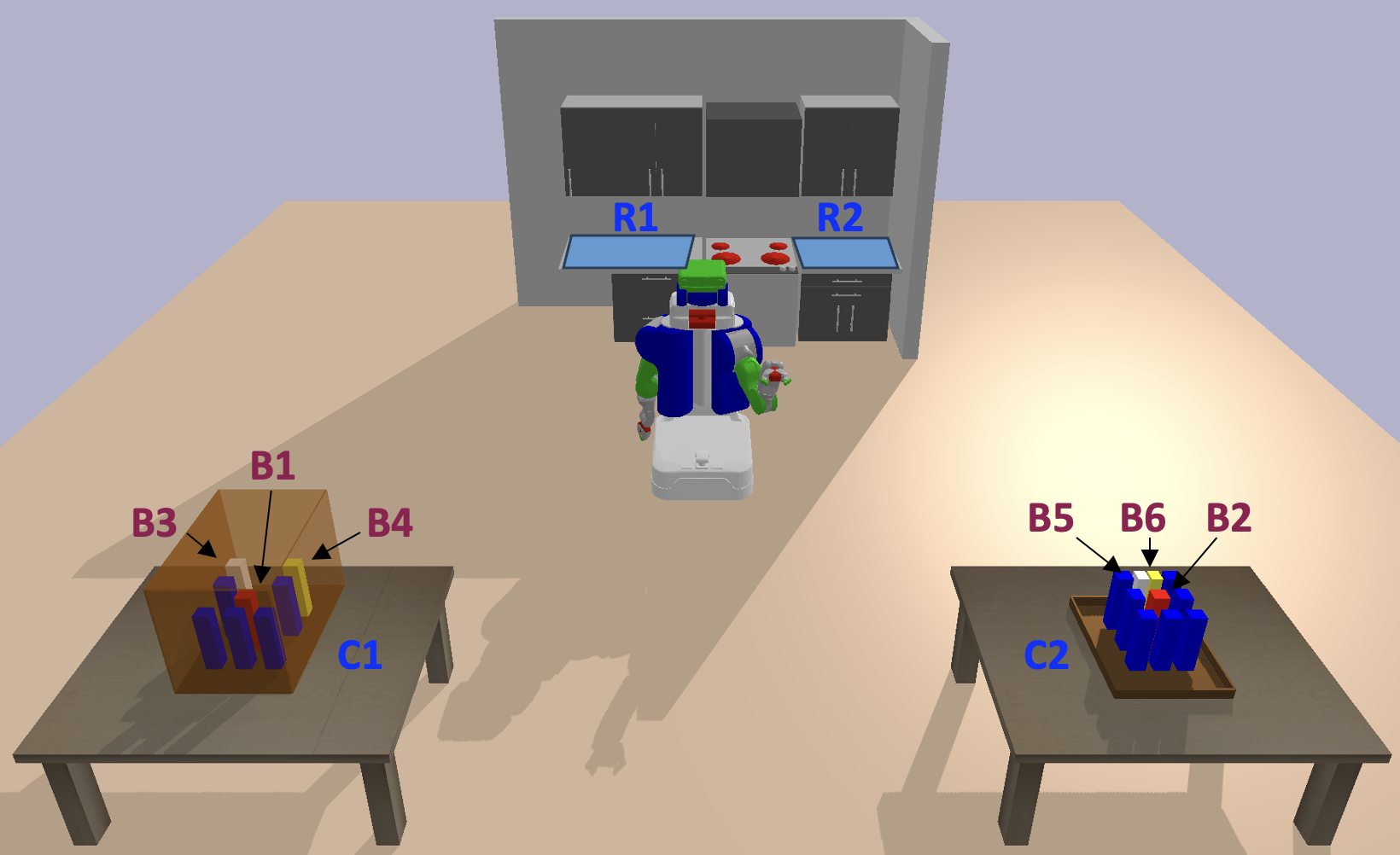}
\caption{Manipulation domain.}
\end{subfigure}
\caption{Visualization of domains used in the example scenarios.}
\label{fig:domain}
\end{figure*}

\subsection{Navigation domain}
\label{subsec:navigation}
The visualization of the navigation domain is included in Figure~\ref{fig:domain} (1), where the target office, start office, and intermediate offices are depicted. The navigation domain considers only fixed objects corresponding to individual office rooms, excluding movable objects, thus $\mathcal{O}=\mathcal{O}_F$. 

We only utilize the mobile base of the PR2 to use it as a mobile robot. Therefore, the configuration space becomes $\mathbb{Q}\in SE(2)$, specifying $(x, y, \theta)$. The planner has access to the xy domain of the workspace of each office room and can determine if the robot is located in the office.

In this domain, four forms of predicates $\mathcal{P}$ are used. Their corresponding literals are as follows: 
\begin{itemize}
\item $\texttt{InRoom}(o_f\in\mathcal{O}_F, q\in\mathbb{Q})$, true if the robot, at configuration $q$, is within the office room $o_f$.
\item $\texttt{Reachable}(q\in\mathbb{Q}, q^\prime\in\mathbb{Q})$, true if a collision-free path between configurations $q$ and $q^\prime$ is found.
\item $\texttt{AtConf}(q\in\mathbb{Q})$, true if the robot is at configuration $q$.
\item $\texttt{CFree}(q\in\mathbb{Q})$, true if the robot at configuration $q$ is collision-free with any fixed objects.
\end{itemize}
$\texttt{InRoom}$, $\texttt{Reachable}$, and $\texttt{CFree}$ are constant predicates, while $\texttt{AtConf}$ is a fluent predicate as its state varies depending on configuration $q$.

Actions $\Delta$ include a single form, which is $\texttt{move}(o_f\in\mathcal{O}_F, o_f^\prime\in\mathcal{O}_F, q\in\mathbb{Q}, q^\prime\in\mathbb{Q})$. The preconditions and effects of $\texttt{move}$ are:
\begin{itemize}
\item $\textsl{pre}$: $\texttt{InRoom}(o_f, q)$, $\texttt{Reachable}(q, q^\prime)$, $\texttt{AtConf}(q)$, $\texttt{CFree}(q)$.
\item $\textsl{eff}$: $\texttt{InRoom}(o_f^\prime, q^\prime)$, $\neg\texttt{AtConf}(q)$, $\texttt{AtConf}(q^\prime)$, $\texttt{CFree}(q^\prime)$.
\end{itemize}
Here, $\neg$ denotes a negation symbol.

Application of $\texttt{move}$ implies that the robot transitions from configuration $q$ in office room $o_f$ to configuration $q^\prime$ in office room $o_f^\prime$, and a collision-free path between configurations $q$ and $q^\prime$ is obtained as a byproduct of evaluating $\texttt{Reachable}$. $q^\prime$ is determined by uniform sampling within the 2D region of office room $o_f^\prime$. We employ RRT-Connect~\cite{kuffner2000rrt} to compute a collision-free path. 

Initial literals $\mathcal{I}$ are $\{\texttt{InRoom}(o_s, q_{init}), \texttt{AtConf}(q_{init}),$ $ \texttt{CFree}(q_{init})\}$, where $o_s$ denotes the start office room and $q_{init}$ is the initial robot configuration. Goal literals $\mathcal{G}$ are $\{\texttt{InRoom}(o_g, q\in\mathbb{Q}), \texttt{CFree}(q\in\mathbb{Q})\}$ for any configuration $q$ where $\texttt{InRoom}$ is true, and $o_g$ denotes the goal office room.

There are four plan skeletons $\Sigma=\{\sigma_1, ..., \sigma_4\}$ as follows:
\begin{itemize}
\item $\sigma_1=\big(\delta_1^1=\texttt{move}(o_f=R_1, o_f^\prime=R_2), \delta_1^2=\texttt{move}(o_f=R_2, o_f^\prime=R_3), \delta_1^3=\texttt{move}(o_f=R_3, o_f^\prime=R_4), \delta_1^4=\texttt{move}(o_f=R_4, o_f^\prime=R_{13}) \big)$,
\item $\sigma_2=\big(\delta_2^1=\texttt{move}(o_f=R_1, o_f^\prime=R_5), \delta_2^2=\texttt{move}(o_f=R_5, o_f^\prime=R_6), \delta_2^3=\texttt{move}(o_f=R_6, o_f^\prime=R_7), \delta_2^4=\texttt{move}(o_f=R_7, o_f^\prime=R_{13}) \big)$,
\item $\sigma_3=\big(\delta_3^1=\texttt{move}(o_f=R_1, o_f^\prime=R_5), \delta_3^2=\texttt{move}(o_f=R_5, o_f^\prime=R_8), \delta_3^3=\texttt{move}(o_f=R_8, o_f^\prime=R_9), \delta_3^4=\texttt{move}(o_f=R_9, o_f^\prime=R_{13}) \big)$,
\item $\sigma_4=\big(\delta_4^1=\texttt{move}(o_f=R_1, o_f^\prime=R_{10}), \delta_4^2=\texttt{move}(o_f=R_{10}, o_f^\prime=R_{11}), \delta_4^3=\texttt{move}(o_f=R_{11}, o_f^\prime=R_{12}), \delta_4^4=\texttt{move}(o_f=R_{12}, o_f^\prime=R_{13}) \big)$.
\end{itemize}
where arguments for typed variables are omitted from actions. Notice that $\sigma_2$ and $\sigma_3$ plan skeletons share the same action $\texttt{move}(o_f=R_1, o_f^\prime=R_5)$.

\subsection{Manipulation domain}
\label{subsec:manipulation}

Unlike the navigation domain, the manipulation domain includes movable objects $\mathcal{O}_M$, as well as three fixed objects $\mathcal{O}_F=\{\mbox{Worktop}, \mbox{Table}, \mbox{Shelf}\}$, as visualized in Figure~\ref{fig:domain} (2). Here, the goal is to move two movable objects, one from the shelf (\ie, B1) and another from the table (\ie, B2), to the worktop. However, other movable objects are located near B1 and B2, potentially resulting in a long planning time for directly retrieving B1 and B2, which may not meet a deadline. Instead, removing nearby movable objects first may be beneficial, as it creates space and shortens the planning time for grasping B1 and B2.

In this domain, the robot's configuration space is $\mathcal{Q}\in SE(2)\times \mathbb{T}^7$, where $SE(2)$ represents the configuration for the mobile base of the PR2, as used in the navigation domain, and $\mathbb{T}^7$ represents the configuration of the $7$ joints of the PR2's left arm (\ie, $7$-dimensional torus). The pose space of each movable object is $\mathcal{P}_i\in SE(3)$, specifying rigid body transformations in 3D space.

More predicates $\mathcal{P}$ and actions $\Delta$ are introduced to describe physical interactions between the robot and the objects. In addition to $\texttt{AtConf}$ introduced in the navigation domain, the literals of the additional predicates are as follows:
\begin{itemize}
\item $\texttt{InRegion}(o_i\in\mathcal{O}_M, o_f\in\mathcal{O}_F, p_i\in\mathbb{P}_i)$, true if the movable object $o_i$ with pose $p_i$ is in the workspace of the fixed object $o_f$.
\item $\texttt{Reachable}(q\in\mathbb{Q}, q^\prime\in\mathbb{Q}, \forall i\ p_i\in\mathbb{P}_i)$, true if a collision-free path between configurations $q$ and $q^\prime$ is found with all movable objects at poses $p_i$.
\item $\texttt{Grasp}(o_i\in\mathcal{O}_M, p_i\in\mathbb{P}_i, g\in\mathbb{G})$, true if the movable object $o_i$ at pose $p_i$ can be grasped by the robot at grasp pose $g$.
\item $\texttt{Kin}(o_i\in\mathcal{O}_M, p_i\in\mathbb{P}_i, g\in\mathbb{G}, q\in\mathbb{Q})$, true if the robot at configuration $q$, with grasp pose $g$, and the movable object $o_i$ at pose $p_i$, satisfies a kinematic constraint.
\item $\texttt{InHand}(o_i\in\mathcal{O}_M, g\in\mathbb{G})$, true if the movable object $o_i$ is stably grasped by the robot with grasp pose $g$.
\item $\texttt{Empty}$, true if the robot's hand is empty.
\item $\texttt{AtPose}(o_i\in\mathcal{O}_M, p_i\in\mathbb{P}_i)$, true if the movable object $o_i$ is at pose $p_i$.
\item $\texttt{CFree}(q\in\mathbb{Q}, \forall i\ p_i\in\mathbb{P}_i)$, true if the robot at configuration $q$ is collision-free with all fixed objects and movable objects at poses $p_i$.
\end{itemize}
$\texttt{InRegion}$, $\texttt{Reachable}$, $\texttt{Grasp}$, $\texttt{Kin}$, and $\texttt{CFree}$ are constant predicates, while $\texttt{InHand}$, $\texttt{Empty}$, $\texttt{AtPose}$, and $\texttt{AtConf}$ are fluent predicates.

The following three forms of actions are used in the manipulation domain:
\begin{itemize}
\item $\texttt{move}(q\in\mathbb{Q}, q^\prime\in\mathbb{Q}, \forall i\ p_i\in\mathbb{P}_i)$
\begin{itemize}
\item $\textsl{pre}$: $\texttt{Reachable}(q, q^\prime, \forall i\ p_i)$, $\texttt{AtConf}(q)$, $\texttt{CFree}(q, \forall i\ p_i)$.
\item $\textsl{eff}$: $\neg\texttt{AtConf}(q)$, $\texttt{AtConf}(q^\prime)$, $\texttt{CFree}(q^\prime, \forall i\ p_i)$.
\end{itemize}

\item $\texttt{pick}(o_i\in\mathcal{O}_M, p_i\in\mathbb{P}_i, g\in\mathbb{G}, q\in\mathbb{Q})$
\begin{itemize}
\item $\textsl{pre}$: $\texttt{AtPose}(o_i, p_i)$, $\texttt{Kin}(o_i, p_i, g, q)$, $\texttt{Empty}$, $\texttt{Grasp}(o_i, p_i, g)$, $\texttt{AtConf}(q)$.
\item $\textsl{eff}$: $\neg\texttt{AtPose}(o_i, p_i)$, $\neg\texttt{Empty}$, $\texttt{InHand}(o_i, g)$.
\end{itemize}

\item $\texttt{place}(o_i\in\mathcal{O}_M, o_f\in\mathcal{O}_F, p_i\in\mathbb{P}_i, g\in\mathbb{G}, q\in\mathbb{Q})$
\begin{itemize}
\item $\textsl{pre}$: $\texttt{Kin}(o_i, p_i, g, q)$, $\texttt{InHand}(o_i, g)$, $\texttt{AtConf}(q)$.
\item $\textsl{eff}$: $\texttt{InRegion}(o_i, o_f, p_i)$, $\texttt{AtPose}(o_i, p_i)$, $\texttt{Empty}$, $\neg\texttt{InHand}(o_i, g)$.
\end{itemize}
\end{itemize}

The \texttt{move} action is similar to the same action in the navigation domain, except that $\texttt{CFree}$ additionally evaluates collisions with objects. RRT-Connect is used to compute a collision-free path. \texttt{pick} and \texttt{place} actions involve evaluating \texttt{Kin} to find inverse kinematic solutions, for which we use IKFast~\cite{diankov2010automated}. The \texttt{pick} action evaluates \texttt{Grasp}, where a grasp is sampled with respect to the tool frame of the PR2. The \texttt{place} action requires sampling a placement pose $p_i$ of the movable object $o_i$ in the workspace space of the fixed object $o_f$, such as the surface of the table.

Initial literals $\mathcal{I}$ are $\{\texttt{AtConf}(q_{init}), \texttt{Empty}, \texttt{CFree}(q_{init}, $ $\forall i\ p_{i, init}), \texttt{InRegion}(B1, C1, p_{1, init}),$ $ \texttt{InRegion}(B2, C2, $ $p_{2, init}), \texttt{InRegion}(B3, C1, p_{3, init}), \texttt{InRegion}(B4, C1, $ $p_{4, init}), \texttt{InRegion}$ $(B5, C2, p_{5, init}), \texttt{InRegion}(B6, C2, $ $p_{6, init}), \texttt{AtPose}(o_i\in\mathcal{O}_M, p_{i, init}\in\mathbb{P}_i), \}$. Goal literals $\mathcal{G}$ are $\{\texttt{InRegion}(B1, R1, p_1\in\mathbb{P}_1), \texttt{InRegion}(B2, R2, p_2\in\mathbb{P}_2)\}$.

Eight plan skeletons, $\Sigma=\{\sigma_1, ..., \sigma_8\}$, are given as input to this domain. We include the details of the plan skeletons in Appendix~\ref{app:mani_plans} due to limited space.

\section{Experiments}
\label{sec:exp}

In the experiments, we primarily evaluate two key questions: (1) How effectively can our method find a deadline-aware executable plan compared to baseline methods? (2) How efficiently can our heuristic schemes determine the computation allocation policy compared to approaches that attempt to solve the proposed MDP, which has been proven to be NP-hard? Both questions assess our method's performance as an effort allocation scheduler and its computational efficiency.

To address these questions, we design several problem instances for case studies to systematically evaluate the performance of the proposed methods, measured by the statistics of rewards collected from episodes. We also propose several baselines for effective performance comparison. The comparative analysis is conducted with all methods aiming to maximize collected rewards given sufficient computation time. This approach allows us to focus on evaluating the maximum capacity of each method without the constraint of limited computation time.

To complement the comparison analysis, we present an analysis of the computation time for the model-based approaches to validate the efficiency improvements introduced by the heuristics compared to MCTS. Lastly, evaluation results on the example scenarios introduced in Section~\ref{sec:example} are shown again under the sufficient computation time regime.

A machine with an Intel Core i7-12700H CPU @ 4.60GHz and 32 GB of memory was used for the experiments. For MCTS, we set the exploration constant $C$ to 0.5. For PPO, we use a codebase from OpenAI Spinning Up~\cite{achiam2018spinning} and include the details of the network architectures for the actor and critic, as well as the hyperparameters, in Appendix~\ref{app:param_ppo}.

\subsection{Baselines}
\label{subsec:baseline}

We design two additional algorithms as baselines to compare with the proposed algorithms. These baselines are simple and thus expected to perform as lower bounds. The first baseline is Round Robin, which deterministically chooses a plan skeleton for refinement at each time step from a pre-specified order among plan skeletons without needing to learn any distributions or repeat problem-solving. Specifically, Round Robin repeatedly selects from $\sigma_1$ to $\sigma_K$. The second baseline is Greedy, which, like model-based methods, needs to learn distributions and always chooses the plan skeleton with the smallest sum of the mean planning and execution times of all involved actions.

\subsection{Comparison analysis}
\label{subsec:comparison}

\begin{figure*}
\label{fig:exp_trees}
\begin{subfigure}[t]{.2\textwidth}
\centering
\includegraphics[width=1.1\textwidth]{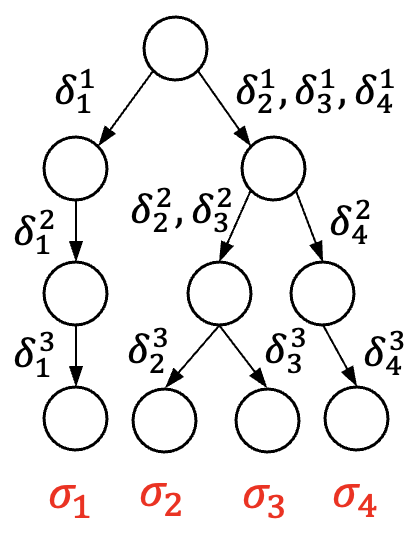}
\caption{Instance 1.}
\end{subfigure}%
\begin{subfigure}[t]{.2\textwidth}
\centering
\includegraphics[width=.76\textwidth]{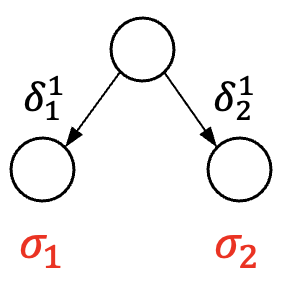}
\caption{Instance 2.}
\end{subfigure}%
\begin{subfigure}[t]{.2\textwidth}
\centering
\includegraphics[width=1.05\textwidth]{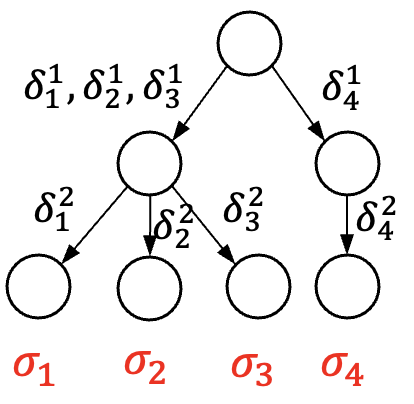}
\caption{Instance 3.}
\end{subfigure}%
\begin{subfigure}[t]{.2\textwidth}
\centering
\includegraphics[width=.76\textwidth]{exp_2.png}
\caption{Instance 4.}
\end{subfigure}%
\begin{subfigure}[t]{.2\textwidth}
\centering
\includegraphics[width=.8\textwidth]{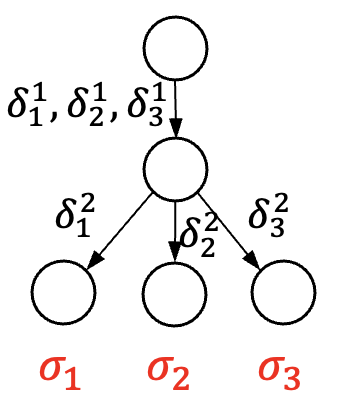}
\caption{Instance 5.}
\end{subfigure}%
\caption{Problem instances designed for the experiments.}
\label{fig:prob_inst}
\end{figure*}

Figure~\ref{fig:prob_inst} depicts five problem instances used for comparison analysis. Instance 1 contains several shared actions, but the mean of planning and execution times is equal for all plan skeletons. Instance 2 also has the same mean for both plan skeletons, and the distributions are symmetric but differ in the variance of planning time distributions. Instance 3 contains a shared action, but the plan skeletons do not have the same mean. Instance 4 is similar to Instance 2 except that the distributions are asymmetric. Instance 5 includes infeasible plan skeletons that cannot be refined even if all available time steps are dedicated. The deadlines pre-specified for these five problem instances are 14, 9, 20, 4, and 14, respectively. Histograms of planning time and execution time for all actions in the five problem instances are provided in Appendix~\ref{app:histogram}.

The performance is measured by the 95\% confidence interval of the return (or cumulative reward) over 100 runs in each problem instance. We do not include the approach of solving the proposed MDP exactly, as its computation time is incomparably greater than all other methods, making it impractical in practice.

\begin{table*}[t]
\begin{center}
\begin{tabular}{ |c||c|c|c|c|c|c| } 
\hline
\hline
\multicolumn{7}{|c|}{Problem instance 1}\\
\hline
Algorithm & MCTS & DP & DP\_Rerun & PPO & Round Robin & Greedy \\ 
\hline
Cumulative reward & $0.98 \pm 0.02$ & $0.93 \pm 0.05$ & $0.94 \pm 0.05$ & $0.72 \pm 0.09$ & $0.37 \pm 0.09$ & $0.92 \pm 0.05$ \\ 
\hline
\hline
\multicolumn{7}{|c|}{Problem instance 2}\\
\hline
Algorithm & MCTS & DP & DP\_Rerun & PPO & Round Robin & Greedy \\ 
\hline
Cumulative reward & $0.74 \pm 0.09$ & $0.65 \pm 0.09$ & $0.70 \pm 0.09$ & $0.37 \pm 0.09$ & $0.51 \pm 0.09$ & $0.65 \pm 0.09$ \\ 
\hline
\hline
\multicolumn{7}{|c|}{Problem instance 3}\\
\hline
Algorithm & MCTS & DP & DP\_Rerun & PPO & Round Robin & Greedy \\ 
\hline
Cumulative reward & $0.71 \pm 0.09$ & $0.35 \pm 0.09$ & $0.67 \pm 0.09$ & $0.25 \pm 0.08$ & $0.59 \pm 0.09$ & $0.34 \pm 0.09$ \\ 
\hline
\hline
\multicolumn{7}{|c|}{Problem instance 4}\\
\hline
Algorithm & MCTS & DP & DP\_Rerun & PPO & Round Robin & Greedy \\ 
\hline
Cumulative reward & $0.94 \pm 0.05$ & $0.90 \pm 0.06$ & $0.90 \pm 0.06$ & $0.79 \pm 0.08$ & $0.45 \pm 0.09$ & $0.54 \pm 0.09$ \\ 
\hline
\hline
\multicolumn{7}{|c|}{Problem instance 5}\\
\hline
Algorithm & MCTS & DP & DP\_Rerun & PPO & Round Robin & Greedy \\ 
\hline
Cumulative reward & $0.71 \pm 0.09$ & $0.33 \pm 0.09$ & $0.70 \pm 0.09$ & $0.10 \pm 0.06$ & $0.70 \pm 0.09$ & $0.34 \pm 0.09$ \\ 
\hline
\hline
\end{tabular}
\end{center}
\caption{Performance comparison among algorithms in five problem instances.}
\label{tab:instance_results}
\end{table*}

Table~\ref{tab:instance_results} shows the results of all methods across five problem instances, each given sufficient computation time to attempt their best performance. MCTS performed the best in all instances, reaching an optimal solution due to the sufficient computation time provided for each iteration of its four phases. DP\_Rerun achieved the second-best performance with only marginal degradation compared to MCTS. This result is surprising, as we will show in the next subsection on the computation analysis between MCTS and DP\_Rerun, where DP\_Rerun required only negligible computation time. The performance of DP was comparable to DP\_Rerun but worse in instances 3 and 5. This was caused by the potential inability to refine the chosen plan skeleton within a deadline, highlighting the importance of rerunning the computation at every time step. From these results, we observe that the contiguity heuristic works effectively in practice, while the  linearity heuristic may face difficulty in some scenarios. 

Greedy performed comparably to the model-based methods in instances 1 and 2 but poorly in the remaining instances. As this method only relies on the mean of distributions, problems like instances 3, 4, and 5 with heavy-tailed distributions that could lead to an unrefinable plan within a deadline can deceive Greedy, underscoring a major limitation of this method.

PPO and Round Robin performed equally poorly. PPO required $10^4$ trials to train the model, which is quite extensive, to achieve this performance. Its computation time was insignificant, as it only required time for the forward pass of the neural network at each time step. While the model-free variant has its value in handling data acquisition issues, dealing with sparse rewards in the presence of deadline constraints remains a significant challenge, left for future work.

Contrary to expectations, Round Robin performed well in instances 3 and 5, likely due to the small number of plan skeletons and deadlines used in the comparison analysis. However, in subsequent analyses involving larger plan skeletons and deadlines in example domains, Round Robin did not succeed even once.

\subsection{Computation time analysis}
\label{subsec:computation}

As metareasoning trades off between solution quality and computation time, better solution quality can be achieved with more computation time. Since the model-free approach inherently differs from the model-based approach by requiring short computation time at the expense of a large number of problem-solving trials, we compare MCTS and DP\_Rerun for this analysis. The more computation time used for MCTS, the closer its solution approaches the optimal solution that can be obtained by solving the MDP exactly. 

\begin{figure}[htbp]
\centering
\includegraphics[width=\linewidth]{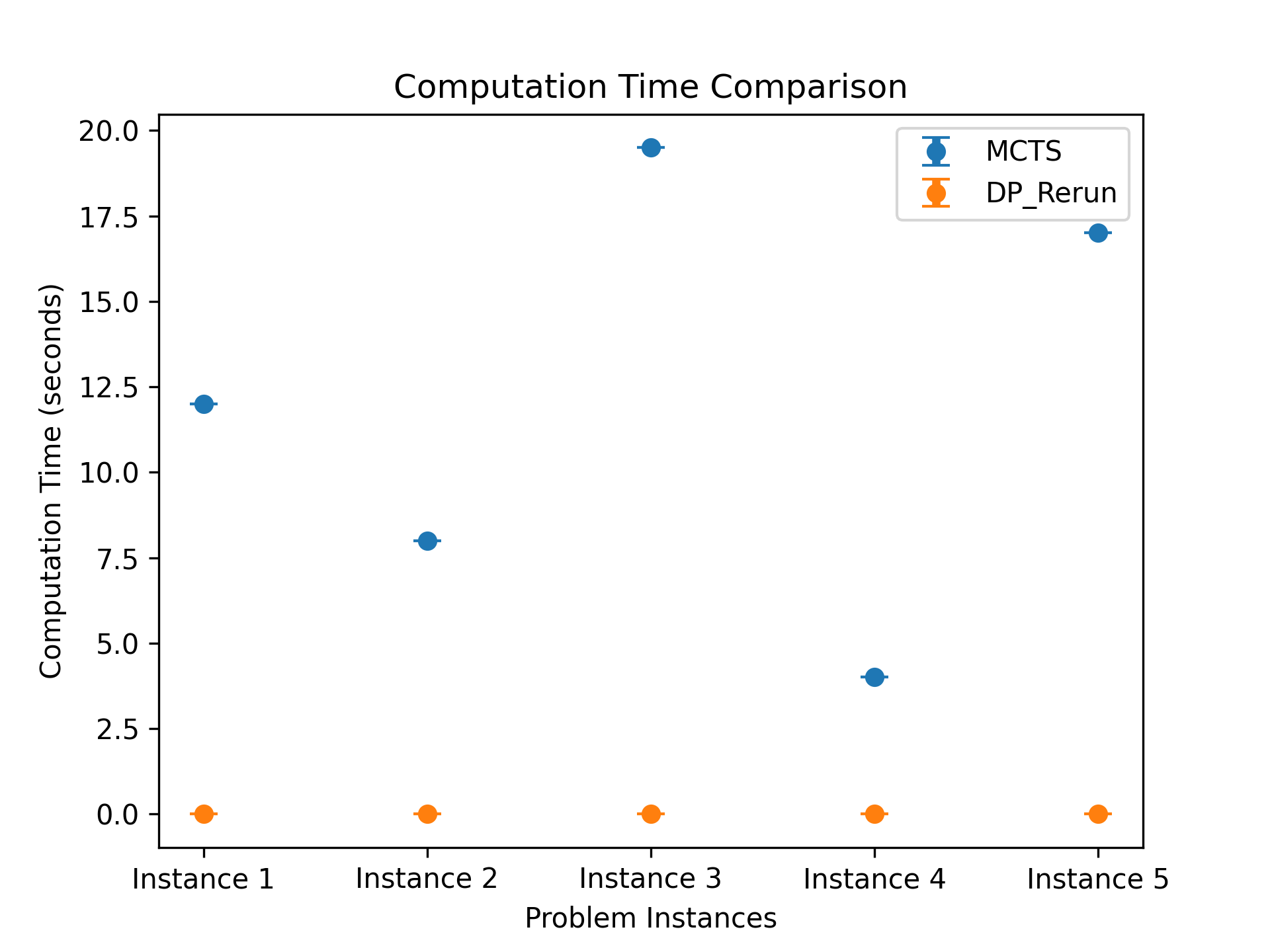}
\caption{Total computation time comparison between MCTS and DP\_Rerun.}
\label{fig:compute_time}
\end{figure}

Figure~\ref{fig:compute_time} shows the total computation times spent solving the five problem instances used in Section~\ref{subsec:comparison}. It can be observed that MCTS required greater computation time to find good-quality solutions, while DP\_Rerun needed negligible computation time to find comparable solutions, demonstrating the dramatic increase in efficiency achieved by the heuristics. 

\begin{figure}[htbp]
\centering
\includegraphics[width=\linewidth]{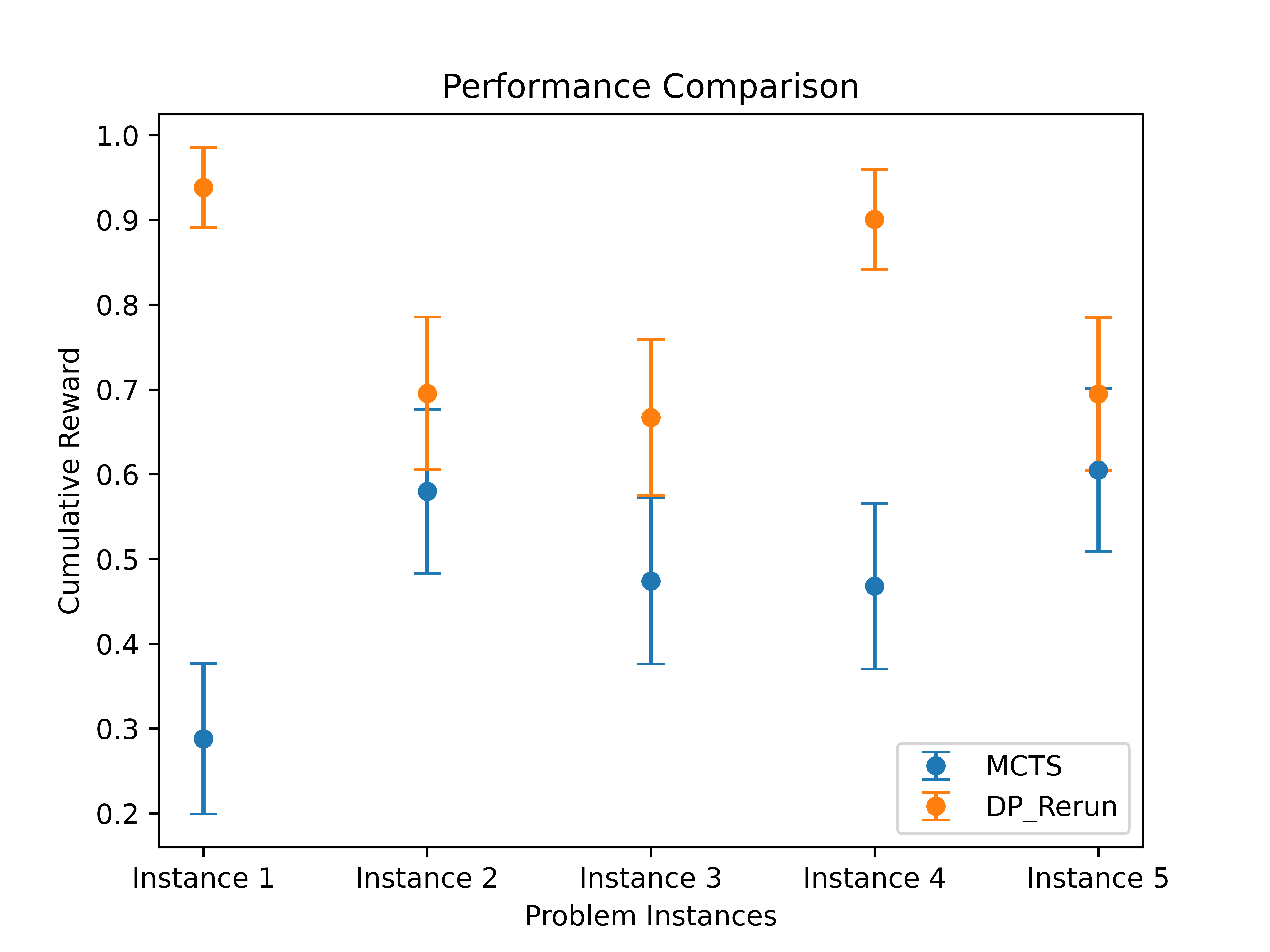}
\caption{Performance comparison between MCTS and DP\_Rerun when both methods are given the same computation time.}
\label{fig:perform_comparison}
\end{figure}

In Figure~\ref{fig:perform_comparison}, we compare the performance, measured by cumulative rewards over 100 runs, between MCTS and DP\_Rerun. Both methods are given the same computation time, which is set equal to DP\_Rerun's computation time. This analysis evaluates the solution quality when both methods are constrained by fixed computation time. The significant degradation in MCTS's performance can be observed, emphasizing DP\_Rerun's ability to maintain reasonable performance even with approximations.

\begin{figure}[htbp]
\centering
\includegraphics[width=\linewidth]{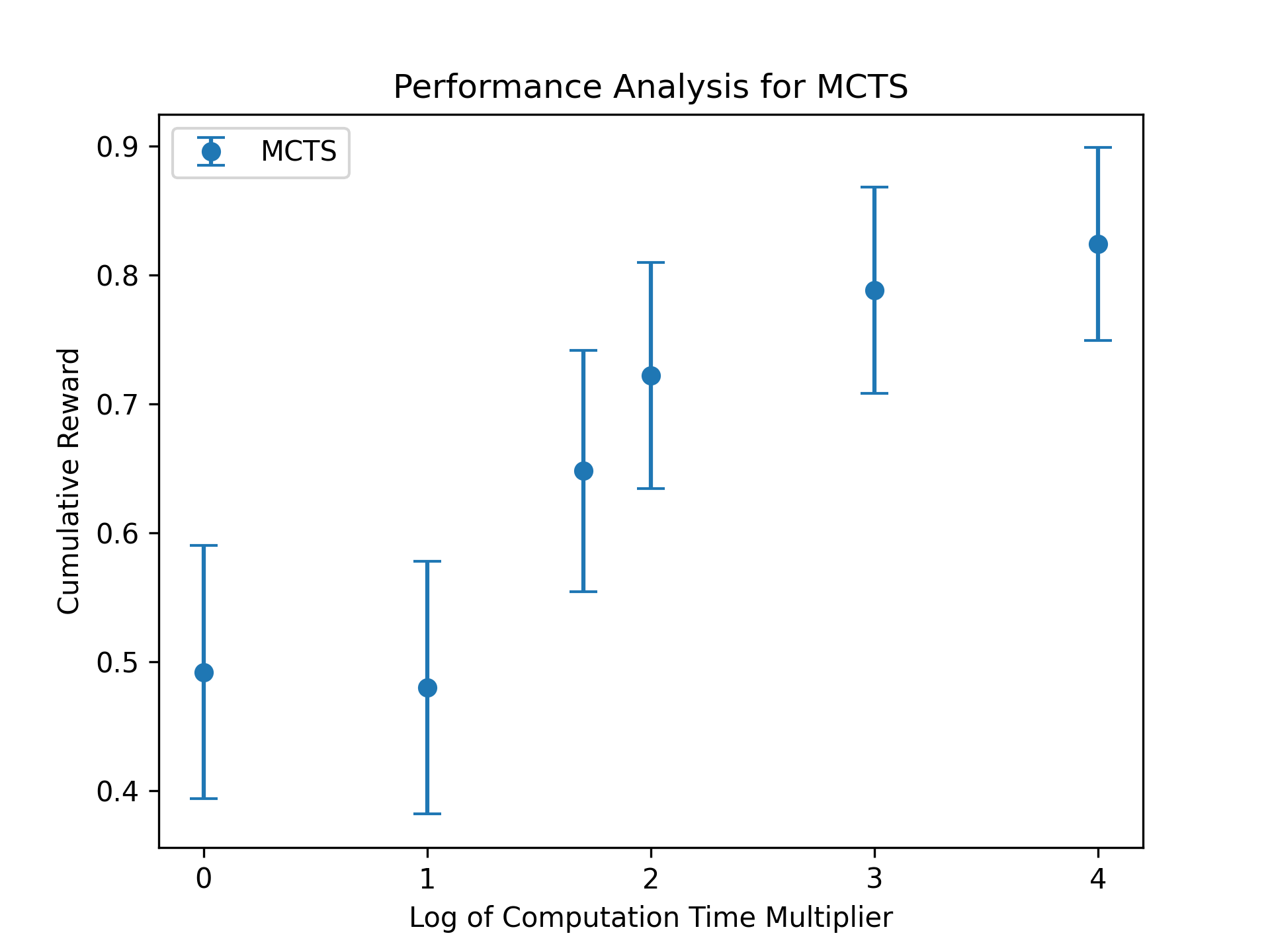}
\caption{Performance analysis of MCTS with varying compuation time.}
\label{fig:mctc_analysis}
\end{figure}

We additionally evaluate the performance of MCTS by varying the computation time across 100 runs of five problem instances. Since each problem instance requires different computation times to find high-quality solutions, we use the logarithm of the computation time multiplier. This multiplier can be applied to compute the actual computation time in seconds for each environment by multiplying a base computation time for DP\_Rerun with ten raised to the power of the multiplier value. The base computation times for DP\_Rerun for each instance are $1.11\times 10^{-5}, 1.26\times 10^{-5}, 2.07\times 10^{-5}, 1.27\times 10^{-5}, 2.12\times 10^{-5}$. More importantly, we can empirically observe the fundamental trade-off in metareasoning, where more computation time leads to better solution quality, in this MCTS performance analysis.

\subsection{Evaluation on example domains}
\label{subsec:eval_domains}
We present the comparison analysis of all the methods in the example domains introduced in Section~\ref{sec:example}. Histograms of planning time and execution time for all actions in both the navigation and manipulation domains are provided in Appendix~\ref{app:histogram}. The performance is again measured by the 95\% confidence interval of the return over 100 runs. The  total decision-making time step $D$ is set to $22$ for navigation and $40$ for manipulation. Like Section~\ref{subsec:comparison}, we provided sufficient computation time for all methods to perform at their best.

\begin{table*}[t]
\begin{center}
\begin{tabular}{ |c||c|c|c|c|c|c| } 
\hline
\hline
\multicolumn{7}{|c|}{Navigation domain}\\
\hline
Algorithm & MCTS & DP & DP\_Rerun & PPO & Round Robin & Greedy \\ 
\hline
Cumulative reward & $0.25 \pm 0.08$ & $0.43 \pm 0.09$ & $0.48 \pm 0.09$ & $0.42 \pm 0.09$ & $0.0 \pm 0.0$ & $0.43 \pm 0.09$ \\ 
\hline
\hline
\multicolumn{7}{|c|}{Manipulation domain}\\
\hline
Algorithm & MCTS & DP & DP\_Rerun & PPO & Round Robin & Greedy \\ 
\hline
Cumulative reward & $0.15 \pm 0.07$ & $0.47 \pm 0.09$ & $0.53 \pm 0.09$ & $0.10 \pm 0.06$ &$0.0 \pm 0.0$  & $0.48 \pm 0.09$ \\ 
\hline
\hline
\end{tabular}
\end{center}
\caption{Performance comparison among algorithms in two example scenarios.}
\label{tab:example_results}
\end{table*}

Table~\ref{tab:example_results} shows the performance results of all methods. In both domains, DP\_Rerun performed the best, followed by DP and Greedy. The improved performances of DP and Greedy compared to those in the five instances are attributed to the tails of distributions not being heavy enough to reduce the probability of plans becoming unrefinable within a deadline, thereby increasing their success rates. Unlike the previous comparison analysis, MCTS performed poorly. This dramatic decrease in performance was due to insufficient computation time, even though it was given significantly more time compared to other methods, averaging 21 seconds for navigation and 45 seconds for manipulation. This result underscores the real complexity of the NP-hard effort allocation problem in large-scale scenarios. PPO was again trained with $10^4$ trials, performing well in navigation but poorly in manipulation due to the large deadline and action space associated with a greater number of plan skeletons.

\section{Conclusion and Future Work}
\label{sec:conc}

This work addresses deadline-aware TAMP problems with the objective of finding a fully executable plan from a set of abstract plans without violating a pre-specified deadline constraint. A metareasoning approach is proposed by formulating an MDP to find optimal effort allocation among abstract plans. Since solving the MDP is proven to be NP-hard, we propose several approximation schemes and further explore a model-free approach that does not require learning the distributions. The proposed methods, particularly DP\_Rerun, show promising results compared to baselines in terms of the cumulative reward related to solution quality and computation time.

As the proposed effort allocation problem introduces a deadline constraint for the first time, this work opens up new avenues for future research. We present several directions that are promising for tackling more complex scenarios and practical applications.

Our TAMP approach can be classified as a sequence-before-satisfy approach introduced in Section~\ref{sec:related}, where abstract plans composed of a sequence of abstract actions are first found, followed by their refinement to create executable low-level motions. If the refinement of any abstract action fails, backtracking is commonly employed to revert to the previously computed low-level motions of the previous abstract action and attempt to find alternative low-level motions. However, the current MDP formulation does not allow for backtracking. If we achieve this capability, then we can expect the effort allocation strategy to effectively handle TAMP problems that may include many infeasible plans where no low-level motions exist.

Another important future direction is to consider various environments where the quantity, class, and shape of objects may vary, instead of a fixed environment. This approach requires representing planning time and execution time distributions that are generalizable to different environments. Generative models may be employed to learn these distributions, which can be conditioned on environment-specific features.

Taking into account exogenous processes in deadline-aware TAMP would allow for addressing a richer class of problems. For example, consider a scenario where one of the abstract actions involves standing in line to grab a coffee, but the waiting time is uncertain. Exogenous processes model uncertain events such as this uncertain waiting time, providing additional sources of uncertainty besides planning and execution times.

The long-term vision of this work is to create scalable, adaptive methods for solving TAMP problems under deadline constraints. As we continue to refine our approach, integrating learning-based methods for generalizing across environments and handling more complex scenarios will be crucial in making deadline-aware TAMP a reliable tool for a wide range of robotic applications.

\section*{Acknowledgments}

\bibliographystyle{IEEEtran}
\bibliography{references}

{\appendices

\input{appendix}

}


 




\vfill

\end{document}

%% file: appendix.tex
\section{Plan Skeletons for the Manipulation Domain}~\label{app:mani_plans}

We show the details of the plan skeletons for the manipulation domain. As numerous actions are involved in this domain, we combine \texttt{move} and \texttt{pick}, as well as \texttt{move} and \texttt{place}, into single abstract actions for simplicity.

\begin{itemize}
\item $\sigma_1=\big(\delta_1^1=\texttt{move}+\texttt{pick}(o_1=B1), \delta_1^2=\texttt{move}+\texttt{place}(o_1=B1, o_f=R1), \delta_1^3=\texttt{move}+\texttt{pick}(o_2=B2), \delta_1^4=\texttt{move}+\texttt{place}(o_2=B2, o_f=R2) \big)$,
\item $\sigma_2=\big(\delta_2^1=\texttt{move}+\texttt{pick}(o_1=B1), \delta_2^2=\texttt{move}+\texttt{place}(o_1=B1, o_f=R1), \delta_2^3=\texttt{move}+\texttt{pick}(o_6=B6), \delta_2^4=\texttt{move}+\texttt{place}(o_6=B6, o_f=C2), \delta_2^5=\texttt{move}+\texttt{pick}(o_2=B2), \delta_2^6=\texttt{move}+\texttt{place}(o_2=B2, o_f=R2) \big)$,
\item $\sigma_3=\big(\delta_3^1=\texttt{move}+\texttt{pick}(o_1=B1), \delta_3^2=\texttt{move}+\texttt{place}(o_1=B1, o_f=R1), \delta_3^3=\texttt{move}+\texttt{pick}(o_5=B5), \delta_3^4=\texttt{move}+\texttt{place}(o_5=B5, o_f=C2), \delta_3^5=\texttt{move}+\texttt{pick}(o_2=B2), \delta_3^6=\texttt{move}+\texttt{place}(o_2=B2, o_f=R2) \big)$,
\item $\sigma_4=\big(\delta_4^1=\texttt{move}+\texttt{pick}(o_1=B1), \delta_4^2=\texttt{move}+\texttt{place}(o_1=B1, o_f=R1), \delta_4^3=\texttt{move}+\texttt{pick}(o_5=B5), \delta_4^4=\texttt{move}+\texttt{place}(o_5=B5, o_f=C2), \delta_4^5=\texttt{move}+\texttt{pick}(o_6=B6), \delta_4^6=\texttt{move}+\texttt{place}(o_6=B6, o_f=C2), \delta_4^7=\texttt{move}+\texttt{pick}(o_2=B2), \delta_4^8=\texttt{move}+\texttt{place}(o_2=B2, o_f=R2) \big)$,
\item $\sigma_5=\big(\delta_5^1=\texttt{move}+\texttt{pick}(o_3=B3), \delta_5^2=\texttt{move}+\texttt{place}(o_3=B3, o_f=C1), \delta_5^3=\texttt{move}+\texttt{pick}(o_4=B4), \delta_5^4=\texttt{move}+\texttt{place}(o_4=B4, o_f=C1), \delta_5^5=\texttt{move}+\texttt{pick}(o_1=B1), \delta_5^6=\texttt{move}+\texttt{place}(o_1=B1, o_f=R1), \delta_5^7=\texttt{move}+\texttt{pick}(o_2=B2), \delta_5^8=\texttt{move}+\texttt{place}(o_2=B2, o_f=R2) \big)$,
\item $\sigma_6=\big(\delta_6^1=\texttt{move}+\texttt{pick}(o_3=B3), \delta_6^2=\texttt{move}+\texttt{place}(o_3=B3, o_f=C1), \delta_6^3=\texttt{move}+\texttt{pick}(o_4=B4), \delta_6^4=\texttt{move}+\texttt{place}(o_4=B4, o_f=C1), \delta_6^5=\texttt{move}+\texttt{pick}(o_1=B1), \delta_6^6=\texttt{move}+\texttt{place}(o_1=B1, o_f=R1), \delta_6^7=\texttt{move}+\texttt{pick}(o_6=B6), \delta_6^8=\texttt{move}+\texttt{place}(o_6=B6, o_f=C2), \delta_6^9=\texttt{move}+\texttt{pick}(o_2=B2), \delta_6^{10}=\texttt{move}+\texttt{place}(o_2=B2, o_f=R2) \big)$,
\item $\sigma_7=\big(\delta_7^1=\texttt{move}+\texttt{pick}(o_3=B3), \delta_7^2=\texttt{move}+\texttt{place}(o_3=B3, o_f=C1), \delta_7^3=\texttt{move}+\texttt{pick}(o_4=B4), \delta_7^4=\texttt{move}+\texttt{place}(o_4=B4, o_f=C1), \delta_7^5=\texttt{move}+\texttt{pick}(o_1=B1), \delta_7^6=\texttt{move}+\texttt{place}(o_1=B1, o_f=R1), \delta_7^7=\texttt{move}+\texttt{pick}(o_5=B5), \delta_7^8=\texttt{move}+\texttt{place}(o_5=B5, o_f=C2), \delta_7^9=\texttt{move}+\texttt{pick}(o_2=B2), \delta_7^{10}=\texttt{move}+\texttt{place}(o_2=B2, o_f=R2) \big)$,
\item $\sigma_8=\big(\delta_8^1=\texttt{move}+\texttt{pick}(o_3=B3), \delta_8^2=\texttt{move}+\texttt{place}(o_3=B3, o_f=C1), \delta_8^3=\texttt{move}+\texttt{pick}(o_4=B4), \delta_8^4=\texttt{move}+\texttt{place}(o_4=B4, o_f=C1), \delta_8^5=\texttt{move}+\texttt{pick}(o_1=B1), \delta_8^6=\texttt{move}+\texttt{place}(o_1=B1, o_f=R1), \delta_8^7=\texttt{move}+\texttt{pick}(o_5=B5), \delta_8^8=\texttt{move}+\texttt{place}(o_5=B5, o_f=C2), \delta_8^9=\texttt{move}+\texttt{pick}(o_6=B6), \delta_8^{10}=\texttt{move}+\texttt{place}(o_6=B6, o_f=C2), \delta_8^{11}=\texttt{move}+\texttt{pick}(o_2=B2), \delta_8^{12}=\texttt{move}+\texttt{place}(o_2=B2, o_f=R2) \big)$.
\end{itemize}

\section{Network Architecture and Hyperparameter Details for PPO}~\label{app:param_ppo}

Two fully-connected neural networks are used separately for the actor and critic, each comprising three hidden layers of 64 neurons connected by Tanh activations. 

The following lists the hyperparameter values we set for the experiments:
\begin{itemize}
\item Clipping parameter: 0.2,
\item Gamma parameter (discount factor): 0.99,
\item Actor learning rate: 0.0003,
\item Critic learning rate: 0.001,
\item Maximum time steps per episode: 1,000,
\item Number of epochs at each iteration: 80,
\item Time steps per epoch: 4,000.
\end{itemize}

\section{A Collection of Histograms for Planning and Execution Times}~\label{app:histogram}

We present a collection of histograms for planning and execution times for all examples and experiments, each based on 1,000 trials.

Due to space limitations, the histograms are provided at the following link: \url{https://drive.google.com/file/d/1BPQeLd7E3QvLuwr7tyvoKnak3ClhQa91/view?usp=sharing}